\documentclass[letterpaper,10pt,journal]{ieeeconf}    

\IEEEoverridecommandlockouts                         

\usepackage{amsmath}
\usepackage{mathtools}
\usepackage{amssymb}
\usepackage{bm}
\usepackage[table]{xcolor}
\usepackage{algorithm}
\usepackage{algpseudocode}
\usepackage{verbatim}
\usepackage{bibentry}
\usepackage{fancyref}
\usepackage[flushleft]{threeparttable}
\usepackage[font=footnotesize]{caption}
\usepackage[font=footnotesize]{subcaption}
\usepackage{multirow}
\usepackage{accents}
\usepackage{tabularx}
\usepackage{empheq}
\usepackage{makecell}
\usepackage{diagbox}
\usepackage{dashbox}%
\usepackage{hyperref}

\newcommand{\err}[1]{\textcolor[rgb]{1,0,0}{#1}}

\DeclarePairedDelimiter\ceil{\lceil}{\rceil}

\title{\LARGE \textbf{
Model Predictive Control for Dynamic Cloth Manipulation: Parameter Learning and Experimental Validation%
}}

\author{Adrià Luque, David Parent, Adrià Colomé, Carlos Ocampo-Martinez\\ and Carme Torras
\thanks{This work was developed in the context of the project CLOTHILDE ("CLOTH manIpulation Learning from DEmonstrations"), which has received funding from the European Research Council (ERC) under the European Union’s Horizon 2020 research and innovation programmee (Advanced Grant agreement No 741930).}
\thanks{All authors are with the Institut de Robòtica i Informàtica Industrial, CSIC-UPC, Barcelona, Spain. {\tt\small \{aluque, dparent, acolome, cocampo, torras\}@iri.upc.edu}.}%
\thanks{C. Ocampo-Martinez is also with the Automatic Control Department, Universitat Politècnica de Catalunya - BarcelonaTECH, Barcelona, Spain. {\tt\small carlos.ocampo@upc.edu}}
}

\begin{document}

\maketitle
\thispagestyle{empty}
\pagestyle{empty}

\begin{abstract}
Robotic cloth manipulation is a relevant challenging problem for autonomous robotic systems. Highly deformable objects as textile items can adopt multiple configurations and shapes during their manipulation. Hence, robots should not only understand the current cloth configuration but also be able to predict the future possible behaviors of the cloth.

This paper addresses the problem of indirectly controlling the configuration of certain points of a textile object, by applying actions on other parts of the object through the use of a Model Predictive Control (MPC) strategy, which also allows to foresee the behavior of indirectly controlled points. The designed controller finds the optimal control signals to attain the desired future target configuration.

The explored scenario in this paper considers tracking a reference trajectory with the lower corners of a square piece of cloth by grasping its upper corners. To do so, we propose and validate a linear cloth model that allows solving the MPC-related optimization problem in real time. Reinforcement Learning (RL) techniques are used to learn the optimal parameters of the proposed cloth model and also to tune the resulting MPC. After obtaining accurate tracking results in simulation, the full control scheme was implemented and executed in a real robot, obtaining accurate tracking even in adverse conditions. While total observed errors reach the 5~cm mark, for a 30$\times$30 cm cloth, an analysis shows the MPC contributes less than 30\% to that value.
\end{abstract}

\section{INTRODUCTION}

Robots have become a key component for increasing the productivity in industry since the 20th century. Furthermore, nowadays robots are starting to be part of domestic environments. In both situations, we can find deformable objects like textiles. Until now, most robotics research has focused on rigid object manipulation. However, textile industry is nowadays encountering major technological difficulties in automating parts of their production processes. A simple task such as taking a cloth garment from a rack and putting it in a box for shipping is still an open problem which the industry is trying to find a solution for. Highly deformable objects represent a major challenge due to their physical properties: their shape and appearance can continuously vary during manipulation. Because of this, it is not enough that the robot understands the cloth current configuration, but it also needs to predict how its action will change the cloth state.

\begin{figure}[t]
    \centering
    \includegraphics[width=0.85\linewidth]{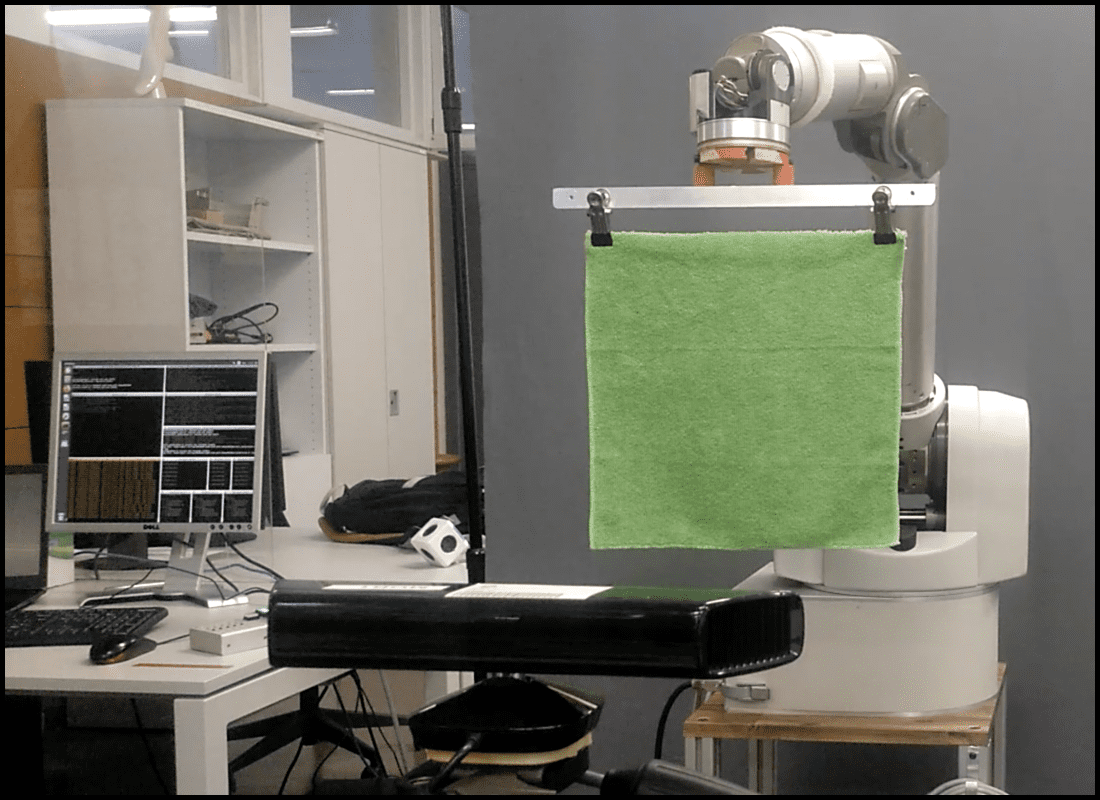}
    \caption{Picture of the setup used in the real-world experiments}
    \vspace{-9mm}
    \label{fig:realsetup_overview}
\end{figure}

In a previous work \cite{SOTA_adria}, Dimensionality Reduction (DR) techniques applied to motion characterization together with Reinforcement Learning (RL) were successful at learning to fold a polo shirt. However, the authors realized how sensitive the output of the action was to any perturbation on the initial conditions. This results in large noise when mapping a robot motion parametrization to the reward function to be optimized in the training phase. Feedforward models have also been used in robotic cloth manipulation \cite{acolome_ICRA2015}, in which the control action results from the sum of the outcome of the Inverse Dynamical Model (IDM) of the robot, i.e., the torque necessary to maintain a tuple of position, velocity and acceleration, and a PID compensating term that accounts for the model deviation from ground truth and external disturbances, such as cloth dynamics. While this kind of controller allows for smaller gains and thus more compliance, it loses precision with larger unmodelled external disturbances. Ideally, one would include the IDM of the manipulated object so as to compensate for its dynamics. While in \cite{ropeIDM}, an IDM model for a rope is learned through CNNs after 60k training interactions, learning such models for a cloth garment would be more sample-expensive and, therefore, it is probably not the best option for cloth manipulation.  Nevertheless, dynamic knowledge of the manipulated object needs to be included in the control loop as, otherwise, the behavior of the cloth may drift away from the desired one, with the controller being unable to compensate such error. This fact hinted the need to include some kind of predictive behavior and prior knowledge in learning cloth manipulation. Therefore, the natural next step is to use a control technique that makes the robot proactively modify its motion according to the predictions made based on a cloth model. 

A suitable technique along this line is Model Predictive Control (MPC) \cite{mpc_rawlingsbook}. The idea of MPC is to use a mathematical model of the plant to be controlled together with an optimization algorithm. This algorithm looks for the best possible control inputs according to a previously defined cost function in a finite time horizon \cite{MPC}. The model is used to predict the future states along that time horizon. In addition, constraints in the related optimization problem can be specified in a quite straightforward way. Bhardwaj et al. \cite{taskMPC} recently used MPC to control a robot under certain restrictions (e.g.: joint limits, collision avoidance, etc.) given its dynamic model. In our case, the plant to be modelled and controlled is a cloth garment. Cloth models are represented as high dimensional systems \cite{SOTA_model1}. These dimensions are required for properly describing cloth behavior, but they make the control problem harder. In literature, different cloth models that simulate the internal dynamics of clothes are available \cite{deformable_model_1987}, \cite{deformable_model_2}. The models consist in solving large systems of equations. However, the high cost of solving these systems limits their utility for real-time applications \cite{cloth_large_eq}. Modeling the dynamics of woven fabrics is a complex problem widely studied in computer graphics. Most models available in literature need to be highly nonlinear to properly describe a realistic behavior \cite{cloth_model}. These nonlinearities will clearly affect the solving speed of an optimization problem. In other works like \cite{SOTA_mpcforce}, the cloth model is included in the MPC design but the solution is obtained after hours of computation.

Cloth manipulation has been a research challenge with successful cases such as the PR2 robot folding towels \cite{berkeley_towel_25min}, where a vision-based grasping point detection algorithm is presented. The robot begins by picking up a dropped towel on a table, goes through a sequence of vision-based re-grasps and manipulations and finally stacks the folded towel in a target location. Despite the impressive results, it takes almost 25 minutes to recognize the different states and complete the task. More recently, Yan et al. \cite{MPClearnModel} used latent representations to learn the dynamic models of non-rigid objects and used them for planning sequences of actions in order to obtain a desired state of the manipulated object. The manipulation is, however, not real-time in the sense that the models learned are used for predicting the outcomes of the actions, rather than real-time control. 
In \cite{SOTA_mpcforce}, the authors presented a technique to synthesize dexterous manipulation of cloth for physics-based computer animation. An optimization problem is formulated to find the commanding forces of the hand so that the cloth follows the reference motion. However, the computational cost is prohibitive for any real-world scenario. Moreover, in 2018 Erickson et al. \cite{SOTA_haptic_MPC} presented an MPC approach that allows a robot to reduce the force exerted during assisted dressing. Nonetheless, the cloth mathematical model was not included in their problem. In the present work, we are interested in incorporating the cloth model in the controller so that the robot always considers the motion of the cloth caused by the movement of its end-effector. The reference motion of the cloth will be represented by the desired motion of certain \textit{interest points}. Up to the knowledge of the authors, MPC of cloth manipulation as we want to tackle it is not found in literature. 

Therefore, the main contribution of this paper is the design, simulation and final implementation on a real setup of a closed-loop control strategy aimed to improve robotic cloth manipulation, by using an MPC that includes a dynamic cloth model satisfying both physical and operational constraints. This controller finds the optimal control inputs (motions of the \textit{grasped points}) that yield minimum tracking error, predicting the behavior of the interest points using a cloth Control-Oriented Model (COM). This model, built and validated with captured evolutions of a real cloth piece, focuses on describing the motion of the aforementioned interest points accurately when given the control signals (motions of the grasped points), in contrast with other cloth dynamic models that try to describe the behavior of the entire piece accurately. Furthermore, RL techniques are used in order to validate the model and find the optimal tuning to reduce tracking error. These contributions are validated in experiments executed in a real robot in real time.

The remainder of this paper is structured as follows: In Section~\ref{sec:problem_statement}, we present the cloth manipulation problem. In Section~\ref{sec:proposed_solution}, we explain our proposed solution: the linear cloth model and the design of the MPC controller. In Section~\ref{sec:CS_description}, we show the case study, specifying differences between simulations and real setup. In Section~\ref{sec:results}, we present our results, including model validation, controller tuning, simulations and trajectory tracking in a real setup. Finally, in Section~\ref{sec:conclusions} we extract conclusions of the achieved results and we explain the forthcoming work directions.

\section{PROBLEM STATEMENT}\label{sec:problem_statement}
\begin{figure}[ht!]
    \centering
    \includegraphics[width=0.75\linewidth]{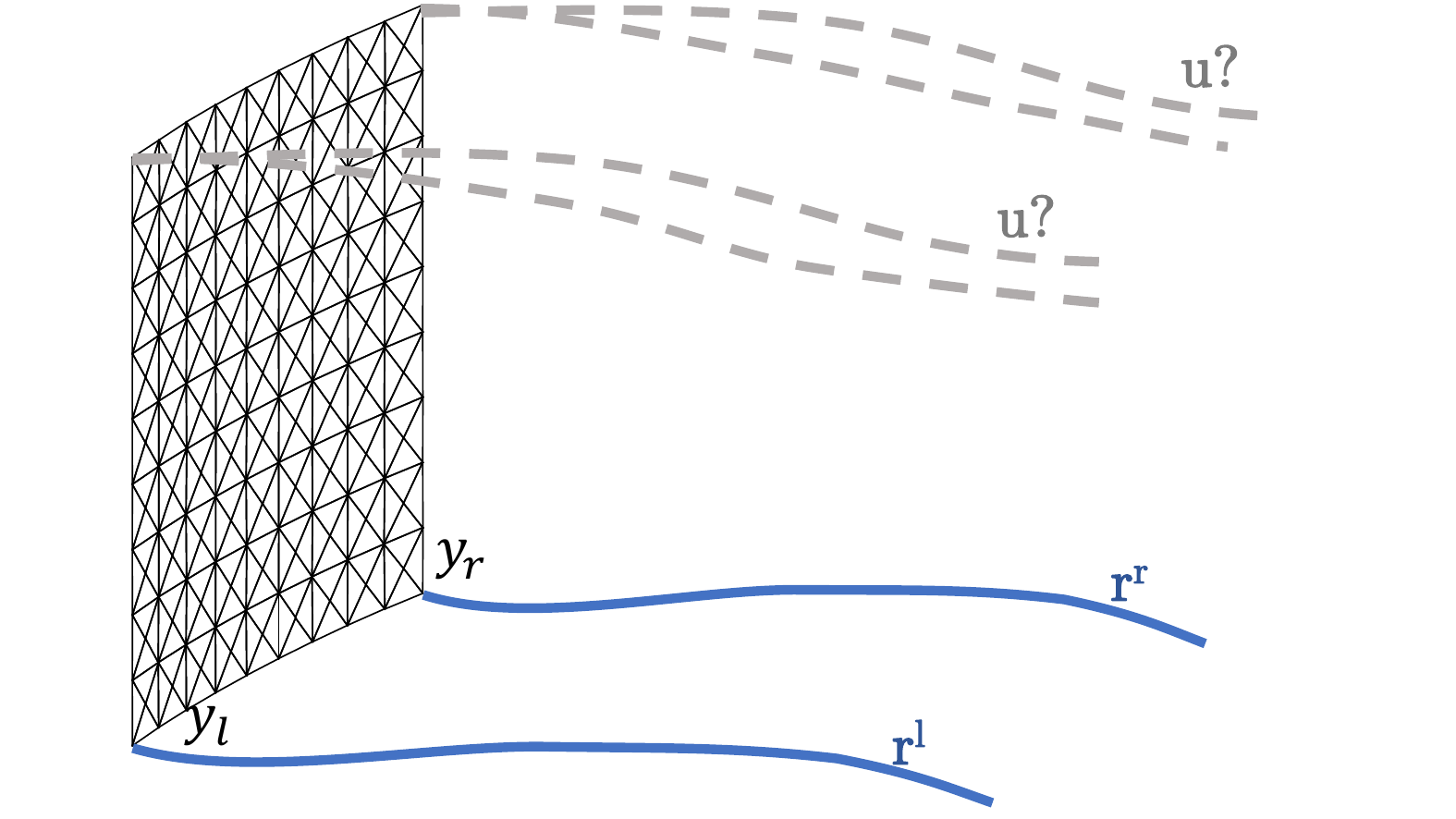}
    \caption{Representation of the trajectory tracking problem}
    \label{fig:problem_statement}
\end{figure}

The main application of this paper is to control the movement of a piece of cloth so that certain parts of such cloth, the \textit{interest points}, track a reference trajectory, by controlling other \textit{grasped points}. In this proof of concept we do not include collision scenarios, leaving this aspect as further work. For this tracking endeavor, we consider that the manipulation is done with two robotic arms by holding the piece of cloth in the air with two pinch grasps. Hence, in our case we manipulate the cloth by varying the position ($x,y,z$ coordinates) of the two upper corners, which are our grasped points $\bm{u} \in\mathbb{U}\subset\mathbb{R}^{6}$  (see Fig. \ref{fig:problem_statement}), where $\mathbb{U}=\{ \bm{u}\in\mathbb{R}^{6} : \underline{\bm{u}}\leq \bm{u}\leq\overline{\bm{u}}  \}$, with $\underline{\bm{u}}$ and $\overline{\bm{u}}$ being the minimum and maximum control signals, respectively. This situation can also be adapted to a single robot arm with a rigid link between the grasped points.

In most manipulation applications, the positions of all the points of the piece of cloth are not relevant for its control, and the efforts can be focused in tracking \textit{interest points}, which have a known relationship with the grasped ones, for example a dynamical model. In our case, defining the trajectory of both lower corners $\bm{y}=\{\bm{y}_r, \bm{y}_l \}^T\in\mathbb{Y}\subset\mathbb{R}^6$ is enough to generate a reference trajectory to track, namely $\bm{r}=\{\bm{r}_r, \bm{r}_l \}^T \in\mathbb{Y}\subset\mathbb{R}^6$ (see Fig. \ref{fig:problem_statement}), where $\mathbb{Y}=\{ \bm{y}\in\mathbb{R}^{6} : \underline{\bm{y}}\leq \bm{y}\leq\overline{\bm{y}}  \}$, with $\underline{\bm{y}}$ and $\overline{\bm{y}}$ being the minimum and maximum position values for each lower corner, respectively. Controlling the position of the lower corners by only moving the upper corners (the grasped ones) is not a simple task, specially when they are far from each other. Dynamic relationships among all the points of the piece of cloth are mostly nonlinear and depend on both the position and velocity of the other points of the cloth.

As stated before, the considered control strategy includes some kind of both system and environment behavioral prediction and prior knowledge of the cloth dynamics. Such a suitable strategy is Model Predictive Control (MPC), where the model is used to predict the future behavior of the system states (and then outputs) along a time-ahead horizon, while physical and operational constraints are satisfied. In the context of this paper, the predictive control strategy requires a mathematical model to predict the cloth behavior. This model must resemble the real cloth statically and dynamically. In literature, we can find multiple theoretical cloth models. Most of them are highly nonlinear and not appropriate for real-time control but suitable for simulation. Therefore, we propose to obtain a linear cloth model, faster but accurate enough, such that the designed MPC strategy can be used in real time, assuming a possible loss of realistic system evolutions in extreme circumstances (e.g., fast dynamics). This reduced model is established and further described in Section~\ref{sec:proposed_solution}. Thus, as previously explained, our idea is to use a realistic Simulation-Oriented Model (SOM) to represent the real cloth in simulation, while the simpler but accurate enough Control-Oriented Model (COM) is to be used for control purposes.

Although the control strategy has certain degree of inherent robustness, and the controller outputs an optimal solution to its optimization problem, its formulation includes several parameters, weights and alternatives that might help reaching minimum tracking errors. To find the best controller structure and its optimal tuning, we propose the usage of RL techniques to learn them automatically.

Summarizing, the problem consists in finding the optimal sequence of control inputs ($\bm{u}$) via MPC to manipulate a cloth, making the interest points ($\bm{y}$) track a desired reference ($\bm{r}$) while satisfying the physical and operational constraints. To this end, we will find a fast COM for real-time implementation, able to reproduce the system dynamics to be controlled, we will use RL to learn its parameters and we will validate the model against the behavior of real cloth pieces. Additionally, the optimal tuning of the controller will also be found with RL to minimize tracking error in closed-loop.

\section{PROPOSED SOLUTION}
\label{sec:proposed_solution}

In this section, we will firstly define the COM to use in our predictive controller in Section~\ref{sec:COM}, followed by the definition of the control optimization problem in Section~\ref{sec:MPCproblem}, with the specific real-time requirements detailed in Section~\ref{sec:RTC}.

\subsection{Control-Oriented Model Definition}
\label{sec:COM}
\begin{figure}[h]
    \centering
    \vspace{-3mm}
    \includegraphics[width=0.27\linewidth]{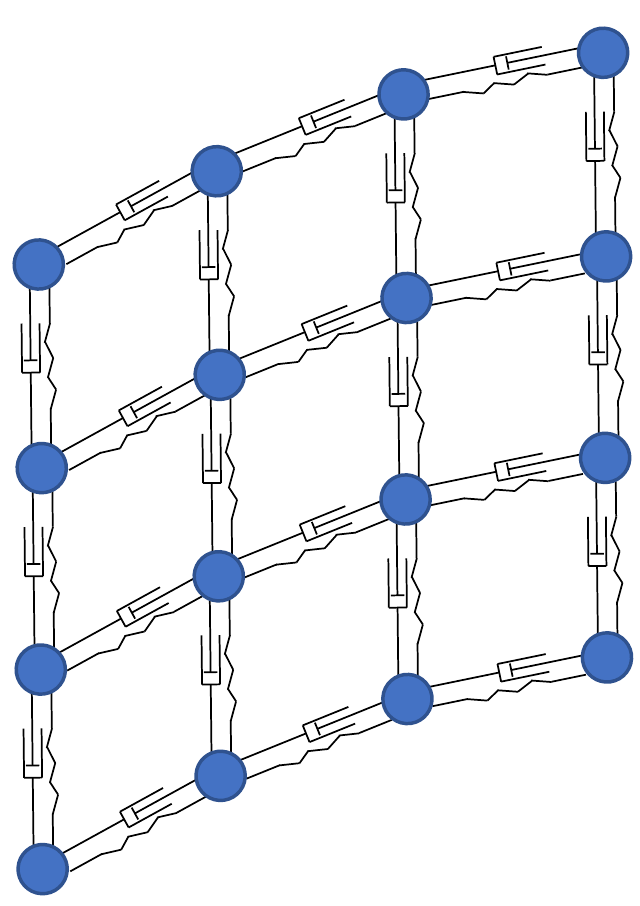}
    \caption{Mass-spring-damper system with structural springs ($N=n^2=16$)}
    \vspace{-3mm}
    \label{fig:springs_mesh}
\end{figure}

We propose to use the simplest --and cheapest to evaluate-- cloth model: a mass-spring-damper system \cite{mass_spring_1}. This model should be able to reproduce those system dynamics to be controlled afterwards, while maintaining highest degree of accuracy with respect to the real cloth. Here, the cloth is treated as a system of particles (nodes) interconnected with spring-dampers. Usually, mass-spring models use three types of connections between nodes to give them more realism \cite{mass_spring_1}: structural springs, shear springs and flexion springs. Our main goal is not the most realistic model but a fast model that properly describes the lower-corners behavior along a prediction horizon $H_p$. Note that, whilst we want such a model to fit the behavior of the lower corners as realistically as possible, the behavior of rest of the mesh is only enforced indirectly. Hence, we are only going to use structural springs as shown in Figure \ref{fig:springs_mesh} with a mesh of $N$ nodes. In a general case, $N=n_r \times n_c$, with $n_r$ and $n_c$ being the number of nodes per row and column, respectively, but in a square mesh, we have $N=n^2$.

The system in Figure 3 can be represented as a graph of $N$ nodes belonging to a set  $\mathcal{I} = \{1,2,...,N\}$ and a set of edges $\mathcal{E}$ that represent the springs connecting those nodes. Moreover, a set of neighbouring nodes related to the $i$-th node is defined as $\mathcal{N}_i = \{j : (i,j) \in \mathcal{E}\}$, where its cardinality is $|\mathcal{N}_i| = \rho$. In our case, we only use structural springs --which yield a sufficiently proper results and save computational costs--, therefore, $\rho=4$ for the interior points of the mesh. For each particle $i\in\mathcal{I}$, we can compute the resulting force of the springs $\bm{S}_k^i$ in time step $k\in \mathbb{Z}_{\geq 0}$ as
\begin{equation*}\label{eq:springforce}
    \bm{S}_k^i = \sum_{\mathcal{N}_i} -K\frac{\|{\bm{d}_k^j} \|-l_0^j}{\|{\bm{d}_k^j} \|}{\bm{d}_k^j},
\end{equation*}
where $K$ is the spring stiffness, ${\bm{d}_k^j}\in\mathbb{R}^{3}$ is the vector pointing from the neighbor $j\in\mathcal{N}_i$ to the particle $i\in\mathcal{I}$ and $l_0^j\in\mathbb{R}_{> 0}$ is the initial length of the spring connecting that $j$ neighbor with the $i$-th particle. The resulting force of the dampers $\bm{D}_k^i$ is calculated as $\bm{D}_k^i = \sum_{\mathcal{N}_i} -b(\bm{v}_k^i-\bm{v}_k^j)$, where $b$ is the damping constant and $\bm{v}_k\in\mathbb{R}^{3}$ is the linear velocity of a particle. Moreover, $i\in\mathcal{I}$ and $j\in\mathcal{N}_i$.

In addition, each particle is affected by the force of gravity $\bm{G}_k^i = -m^i \bm{g}$, where $m^i$ is the mass of the particle $i\in\mathcal{I}$ and $\bm{g}$ is the gravity constant. By adding these three terms, we compute the total force $\bm{F}_k^i$ applied to the particle as $\bm{F}_k^i~=~\bm{S}_k^i + \bm{D}_k^i + \bm{G}_k^i$. Once we have computed all the forces, we can use Newton's Second Law to compute the acceleration as $\bm{a}_k^i = {\bm{F}_k^i}/{m^i}$.

Having the accelerations of the nodes, we can integrate them over time to obtain the positions and velocities. In the next subsections, we present the proposed steps to obtain a reliable linear cloth model from the accelerations previously calculated.

\subsubsection{Integration of the model}
To integrate the acceleration and obtain the position and velocity of the nodes, we can use either implicit or explicit integration methods. The former are stable but require solving large systems of equations \cite{cloth_large_eq}. The high cost of solving these systems limits their utility for real-time applications. Instead, explicit integration methods are fast but could have stability problems. To avoid these problems, we can put enough damping $\bm{D}^i_k$ in the system so that the energy decreases in a single time step \cite{cloth_stable}. We propose to use explicit Euler's integration method with a given time step $\Delta t$ to update positions and velocities.

\subsubsection{Linearization of the model}
Even if we have proposed a fast spring-damper model, the spring force vector $\bm{S}^i_k$ is nonlinear since it depends on a quadratic norm of state variables multiplied by other state variables. This norm computes the spring length in three dimensions along $k\in \mathbb{Z}_{\geq 0}$ as $\|{\bm{d}_k^j} \| = \sqrt{(p_{x,k}^i-p_{x,k}^j)^2+(p_{y,k}^i-p_{y,k}^j)^2+(p_{z,k}^i-p_{z,k}^j)^2}$. Instead of computing the $\mathcal{L}$2-norm, we propose to calculate it as an $\mathcal{L}1$-norm to linearize it by creating three linear springs, one for each direction of the space. Using this norm, we propose to associate different stiffness constants for each direction $k_x,k_y,k_z$ so as to preserve the different behavior in each direction, i.e.,
\begin{equation*}
    \bm{S}_k^i 	\approx  \sum_{\mathcal{N}_i}
    \begin{bmatrix}
    -k_x ( p_{x,k}^i - p_{x,k}^j - l_{0x}^j) \\
    -k_y ( p_{y,k}^i - p_{y,k}^j - l_{0y}^j) \\
    -k_z ( p_{z,k}^i - p_{z,k}^j - l_{0z}^j) 
    \end{bmatrix}.
\end{equation*}

With this transformation, we have a linear system that can be expressed through a state-space realization as
\begin{align}\label{eq:state_space_model}
\begin{split}
    \bm{x}_{k+1} &= \bm{A} \bm{x}_k + \bm{B} \bm{u}_k + \bm{f}_{ct}, \\
    \bm{y}_{k}   &= \bm{C} \bm{x}_k ,
\end{split}
\end{align}
where $\bm{x}\in\mathbb{X}\subset\mathbb{R}^{6N}$ is the vector of the cloth states (position $p$ and velocity $v$ in $x,y,z$ of nodes $i\in\mathcal{I}$), $\bm{u}\in\mathbb{U}\subset\mathbb{R}^{6}$ is the control input vector, $\bm{y}\in\mathbb{Y}\subset\mathbb{R}^{6}$ is the output vector (positions of the lower corners) and $\bm{f}_{ct}\in\mathbb{F}\subset\mathbb{R}^{6N}$ is the vector of constant forces applied to each node: gravity and natural spring length force. Moreover, $\bm{A}\in\mathbb{R}^{6N\times 6N}$, $\bm{B}\in~\mathbb{R}^{6N\times 6}$, $\bm{C}\in\mathbb{R}^{6\times 6N}$ are the system state-space matrices and
$\mathbb{X}=\{ \bm{x}\in\mathbb{R}^{6N} : \underline{p}_x\leq p_x \leq\overline{p}_x,\  \underline{p}_y\leq p_y \leq\overline{p}_y,\  \underline{p}_z\leq p_z \leq\overline{p}_z,\  \underline{v}_x\leq v_x \leq\overline{v}_x,\  \underline{v}_y\leq v_y \leq\overline{v}_y,\  \underline{v}_z\leq v_z \leq\overline{v}_z  \}$, where $\underline{p}_i$ and $\overline{p}_i$ represent the minimum and maximum position values for each direction, respectively, whereas $\underline{v}_i$ and $\overline{v}_i$ are the corresponding minimum and maximum velocities.

We previously discussed that the addition of the damping term guarantees the stability of the system: If we have different spring constants for each direction, we also need different damping constants to reduce the energy in each direction. Similarly as before, we compute $\bm{D}^i_k$ as
\begin{equation*}
    \bm{D}^i_k 	\approx \sum_{\mathcal{N}_i}
    \begin{bmatrix}
    b_x ( v_{x,k}^i - v_{x,k}^j) \\
    b_y ( v_{y,k}^i - v_{y,k}^j) \\
    b_z ( v_{z,k}^i - v_{z,k}^j) 
    \end{bmatrix}.
\end{equation*}

\subsubsection{The super-elastic problem}
Elasticity is the major drawback of the mass-spring cloth model, as it might stretch under its own weight \cite{mass_spring_1}. Two solutions are found in the literature: making the springs stiffer or applying a maximum deformation rate. The former has its limits, as it can unstabilize the model. With the latter, the elongation of the springs is corrected if it is over a 10\% of the initial length, making the model nonlinear, and thus the computations slower.

We propose a third option to solve the super-elastic problem: shortening the initial length of the linear springs in the vertical direction in simulation by $\Delta l_{0z}$. When gravity is then applied, it results in a force equilibrium in the vertical direction, avoiding the mesh stretching under its own weight.

\begin{figure}[!ht]
    \centering
    \includegraphics[width=0.78\linewidth]{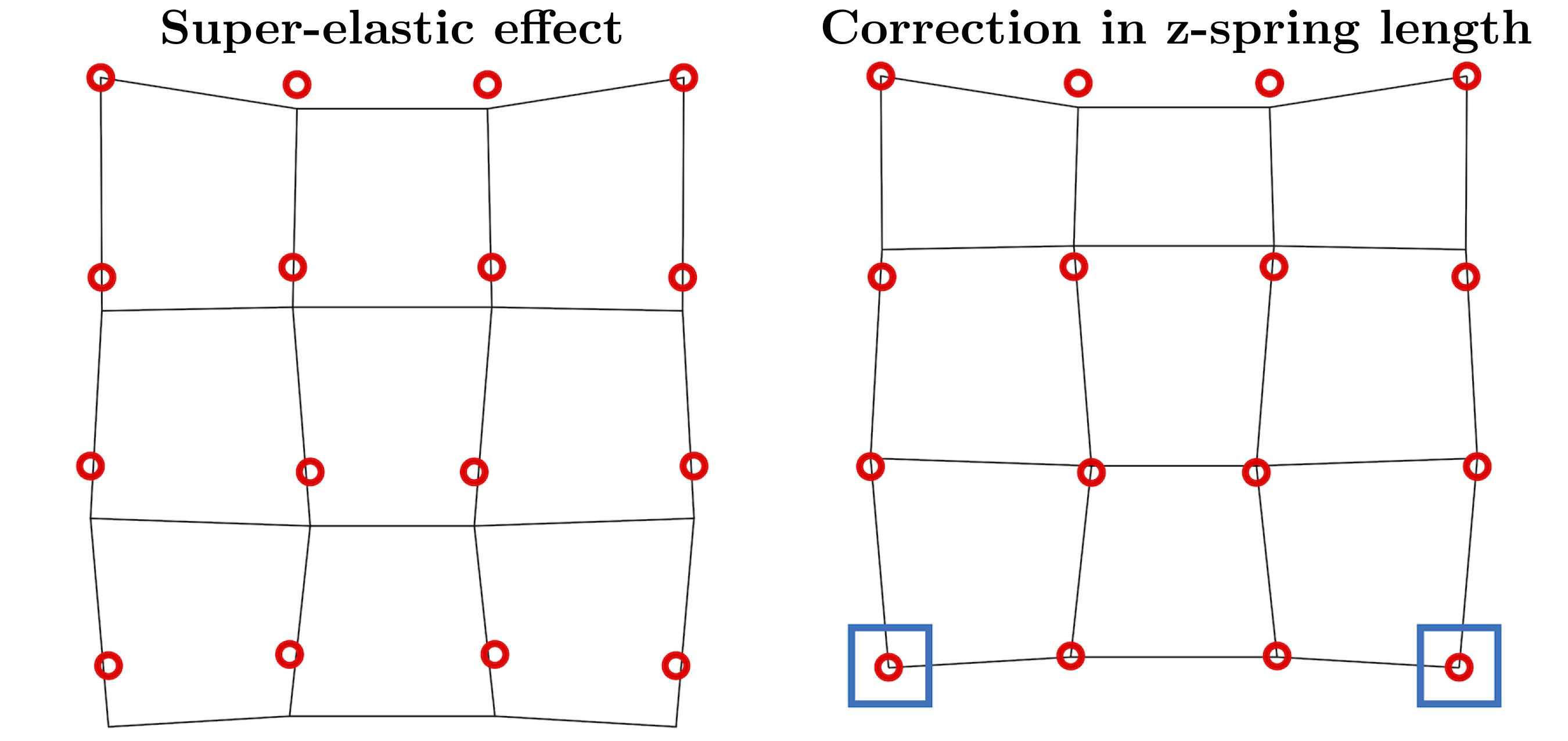}
    \caption{Mesh positions showing the super-elastic problem (left) and the result of applying $\Delta l_{0z}$, correcting it (right)}
    \label{fig:superelastic}
\end{figure}

\subsubsection{Cloth model orientation}
\begin{figure}[!ht]
    \centering
    \includegraphics[width=0.95\linewidth]{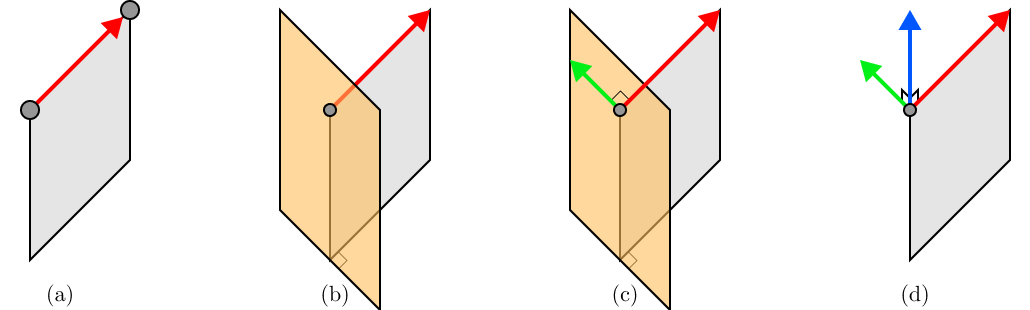}
    \caption{Process to obtain the local cloth base}
    \label{fig:2_rot_params_yes}
\end{figure}

The defined parameters depend on the space axes, and must always be aligned with the cloth model. Therefore, a local reference frame ($L$) must be defined for the cloth model, such that if it rotates, the parameters are expressed in this local base and rotate with it. With the available data during an execution, the fastest approach is shown in Figure \nolinebreak \ref{fig:2_rot_params_yes}.

The first step (Figure \nolinebreak \ref{fig:2_rot_params_yes}(a)) is to define the local $X$ vector (red) in the direction from one corner to the other, $X_L~=~\bm{u}_r-\bm{u}_l$. The other two axes must be contained in the plane perpendicular to this axis, as shown in Figure \nolinebreak \ref{fig:2_rot_params_yes}(b). Without using any other data (full mesh or reference), we can force the local $Y_L$ axis (green) to have null vertical (absolute $Z$) component, as shown in Figure \nolinebreak \ref{fig:2_rot_params_yes}(c), which also defines the local $Z_L$ axis (blue) as the cross product between the other two, $Z_L=X_L\times Y_L$, as seen in Figure \nolinebreak \ref{fig:2_rot_params_yes}(d). This frame definition is suitable for rotations along the $Z_L$ and $Y_L$ axis, but rotations along $X_L$ without changing the positions of the upper corners are ignored, always placing $Y_L$ horizontally.

\subsection{Control Strategy}
\label{sec:MPCproblem}
The main objective of a tracking MPC strategy is to minimize the tracking error while satisfying the system constraints. Closed-loop stability can be guaranteed if the initial state is inside the feasibility region $\bm{x}_0\in\mathbb{X}$ and the evolution of the state trajectories remain inside the polytope defined by the system/operational constraints. The controller has been obtained considering the following elements:

\subsubsection{Cost function}
The goal is to find a sequence control inputs 
\begin{equation} \label{eq:u_seq}
\bm{u}_k^s \triangleq \{ \bm{u}_0,\dots,\bm{u}_{H_p-1}\}    
\end{equation}
along a prediction horizon $H_p$ such that the lower right and left nodes $\bm{y}\in\mathbb{Y}$ follow the desired trajectories $\bm{r}\in \mathbb{Y}$ as accurately as possible. Moreover, we also want to obtain smooth trajectories, reducing noise and sudden changes in acceleration. To this end, reducing the changes in consecutive control signals or \textit{slew rates} (i.e., $\bm{\Delta u}_i~=~\bm{u}_i-\bm{u}_{i-1}$) instead of the control inputs has been proven to yield better results (Section \ref{sec:results}). The final multi-objective cost function at used at each time step $k\in \mathbb{Z}_{\geq 0}$ is

\begin{equation}
    J_k = \sum_{i=0}^{H_p-1} \|\bm{y}_{k+i+1|k}-\bm{r}_{k+i+1}\|^2_{\bm{Q}} + \|\bm{\Delta u}_{k+i|k}\|^2_{\bm{R}},
\label{eq:objfun_Du}
\end{equation}
where the notation $(k+i|k)$ refers to the prediction at time step $(k+i)$ based on measurements made at time step $k$. For $i=0$, $\bm{\Delta u}_{k|k} = \bm{u}_{k|k}-\bm{u}_{k-1|k}$. This $\bm{u}_{k-1|k}=\bm{u}_{k-1}$ is the control input applied in the previous step, and must be given as an initial condition. Equation \eqref{eq:objfun_Du} will be optimized to find $\bm{\Delta u}_{k|k}$, and $\bm{u}_{k|k} = \bm{\Delta u}_{k|k}-\bm{u}_{k-1|k}$ Note that $(\bm{Q},\bm{R})\in\mathbb{R}^{6\times6}$ are the weighting matrices of suitable dimensions that prioritize each term of the multi-objective cost function.

\subsubsection{System constraints}
The robot arms that manipulate the cloth have restrictions of velocity $\bm{v}_{max}$ and workspace $\mathbb{U}$, providing bounds for the search of the optimal computed control input.

\subsubsection{Optimization problem formulation}
Based on the receding horizon strategy \cite{MPC} and considering a fixed prediction horizon $H_p$, the goal is to obtain the input sequences that minimize the tracking errors of the two lower corners while satisfying all constraints. The controller is based on the solution of the following discrete-time open-loop optimization problem (OOP):
\begin{eqnarray}
 & \displaystyle\min_{\bm{u}_k^s} \ J_k, \nonumber \\
 \text{s.t.} &\nonumber \\
& \begin{array}{llc}
  \begin{aligned}
        \bm{x}_{k+i+1|k} &= \bm{A}\bm{x}_{k+i|k}+\bm{B}\bm{u}_{k+i|k}+\bm{f}_{ct} \\
        \bm{y}_{k+i|k} &= \bm{C}\bm{x}_{k+i|k} \\
        \bm{x}_{k|k} &= \bm{x}_k \\
        \bm{u}_{k-1|k} &= \bm{u}_{k-1} \\
        \bm{x}_{k+i+1|k} &\in \mathbb{X} \subseteq \mathbb{R}^n \\
        \bm{u}_{k+i|k} &\in \mathbb{U} \subseteq \mathbb{R}^m, \\
    \end{aligned}  
\end{array}
\label{eq:optprob_slewrate}
\end{eqnarray}
$\forall i\in [0, H_p-1]$, where $\bm{u}_k^s$ is the sequence of control signals in \eqref{eq:u_seq}.

Assuming the OOP \eqref{eq:optprob_slewrate} is feasible, there exists an optimal sequence solution $\bm{u}_k^{s,*}~\triangleq~\{\bm{u}^*_{k|k}, \bm{u}^*_{k+1|k}, ..., \bm{u}^*_{k+H_p-1|k}\}$. The applied control is  $\bm{u}^{MPC}_k \triangleq \bm{u}^*_{k|k},\ (i=0)$, ignoring the rest of the sequence. The entire process is repeated at the next time instant $k$.

\subsubsection{Weighting matrix tuning}
\label{subsect:adaptive}
The $\bm{Q}$ matrix in \eqref{eq:objfun_Du} is the responsible of penalizing the tracking errors for each coordinate of the lower corners, while $\bm{R}$ penalizes sudden changes and quick fluctuations in the control signals. 

An adaptive tuning of the $\bm{Q}$ matrix is proposed. In each time step, we compute the distance vector between the current position and that of the reference at the end of the prediction horizon, finding the main direction $\bm{\beta}$ in which the lower corners move along $H_p$. By normalizing this vector, the final adaptive weighting matrix $\bm{Q}_a$ is computed as
\begin{equation}
    \bm{\beta}_k = \frac{| \bm{r}_{k+H_p}-\bm{y}_{k|k} |}{\| \bm{r}_{k+H_p}-\bm{y}_{k|k} \| + \epsilon},
    \quad \bm{Q}_{a,k} = \mbox{diag}\begin{bmatrix} \bm{\beta}_k \end{bmatrix},
\label{adaptive_Q}
\end{equation}
where an $\epsilon$ term has been included to avoid numerical problems when the corner is exactly at the reference. We must note that this adaptive term changes in each time step, as is made explicit by the $k$ subscript. 

This adaptive computation has been studied as a possible way to compute the optimal tuning along with different constant matrix structures. The results of this study are shown in Section \ref{sec:results}.

\subsection{Real-Time Control}
\label{sec:RTC}

To work in real time, the controller needs to have a constant output rate with an updated control signal on every time step (sampling time $T_s$). It was observed that sometimes, depending on initial conditions, optimizations could take longer than the maximum allowed time (one step), thus slowing down the executions. To avoid this, the solver was changed to run in parallel, and return not only $\bm{u}^*_{k|k}$ but the entire sequence $\bm{u}_k^{s,*}$ of $H_p$ control signals. This way, the solver can be called with new initial conditions on every time step, but if an optimization takes longer than $T_s$ to finish, the controller can output a sub-optimal solution coming from the most recently obtained control sequence. This can be done for a maximum of $H_p$ steps, the length of each sequence, thus a hard time limit was added to cancel optimizations running for too long. In the end, a maximum time of $T_s H_p/4$ was set empirically, ensuring there are always at least four optimizations with different initial conditions in one prediction horizon.

\section{CASE STUDY DESCRIPTION}
\label{sec:CS_description}

This section starts with the specific changes that had to be incorporated to the previously defined solution for the used setup with only one robotic arm, described in Section~\ref{sec:CS_changes}. Then we divide the study in two, with simulations in Section~\ref{sec:CS_sim} and the real setup in Section~\ref{sec:CS_real}.

\subsection{Application to Single-Arm Manipulation}
\label{sec:CS_changes}
The presented formulation in \eqref{eq:optprob_slewrate} assumes the evolutions of the two controlled corners are independent, as they are when using two different robot arms to pick them. In the studied case, only one robot is used, and thus the upper corners are linked together with a rigid piece. This means they always keep a constant distance, adding a new quadratic constraint to the problem. This changes it from Linear to a Quadratically Constrained Quadratic Program (QCQP), but even if it is more complex, convexity is kept \cite{mpc_convex_qcqp}, and computations can be fast. The new OOP is:
\begin{eqnarray}
& \displaystyle\min_{\bm{u}_k^s} \ J_k,  \\
&\begin{array}{llc}
\text{s.t.} & \nonumber \\
&  \begin{aligned}
        \bm{x}_{k+i+1|k} &= \bm{A}\bm{x}_{k+i|k}+\bm{B}\bm{u}_{k+i|k}+\bm{f}_{ct} \\
        \bm{y}_{k+i|k} &= \bm{C}\bm{x}_{k+i|k} \\
        \left\| d^{uc}_{k+i+1|k} \right\|^2 &= L^2 \\
        \bm{x}_{k|k} &= \bm{x}_k \\
        \bm{u}_{k-1|k} &= \bm{u}_{k-1} \\
        \bm{x}_{k+i+1|k} &\in \mathbb{X} \subseteq \mathbb{R}^n \\
        \bm{u}_{k+i|k} &\in \mathbb{U} \subseteq \mathbb{R}^m, \\
    \end{aligned}
\end{array}
\label{eq:optprob_onearm}
\end{eqnarray}
$\forall i\in [0, H_p-1]$, where $d^{uc}$ is the distance vector between the upper corners.

The control signals obtained with the described MPC are the displacements of the upper corners of the cloth in one time step. However, to move one robot, we need a pose of its Tool Center Point (TCP), which must be computed with the available data.  The position of the TCP can be obtained by taking the absolute positions of the upper corners of the cloth, $\bm{u}^{abs}$, computing the midpoint and adding a constant offset ($\Delta h$) introduced by the rigid piece that links both corners together. This offset is always in the $Z_L$ axis of the local cloth base, so it must be transformed to global coordinates before being added as follows:
\begin{equation}
\begin{aligned}
    \Delta \bm{p}^{L} &= \begin{bmatrix}
        0\\0\\ \Delta h
    \end{bmatrix} \rightarrow \Delta \bm{p} = \bm{R}_L^W \cdot \Delta \bm{p}^L, \\
    \bm{p}_{TCP} &= \frac{1}{2} \begin{bmatrix}
        u^{abs}_1 + u^{abs}_2\\
        u^{abs}_3 + u^{abs}_4\\
        u^{abs}_5 + u^{abs}_6
    \end{bmatrix} + \Delta \bm{p}.
\end{aligned}
\label{eq:2_localoffset}
\end{equation}
\textcolor{red}{}
The orientation of the TCP is obtained with the same local reference frame, but inverting the $Z_L$ axis to point away from the robot and into the cloth, as it is a convention in robotics. The other two axes are swapped for practical convenience, resulting in $R_{TCP}^W=[Y_L \ X_L \ \mbox{-}Z_L]$.

\subsection{Simulation}
\label{sec:CS_sim}

\begin{figure*}[htb!]
    \centering
    \includegraphics[width=0.82\textwidth]{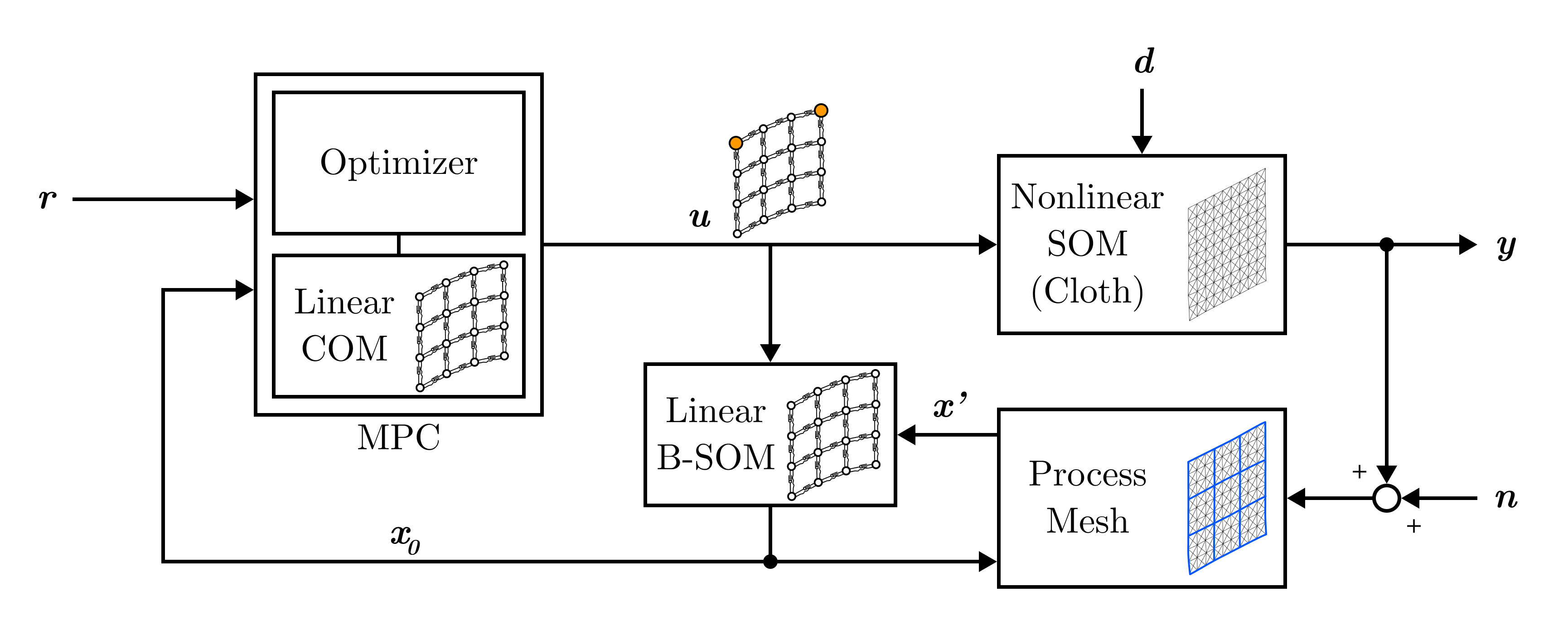}
    \vspace{-1mm}
    \caption{Closed-loop control scheme. The nonlinear SOM substitutes the real cloth in simulation, and the linear B-SOM acts as a fast backup while the real feedback is processed, which in the real setup can take several time steps}
    \vspace{-3mm}
    \label{fig:control_scheme}
\end{figure*}

The proposed control scheme is shown in Figure \ref{fig:control_scheme}. In simulation, instead of using only two models, one inside the controller and one to simulate the real cloth, we add a third model as an intermediate step. This is motivated by the real setup, where the feedback signal from the real cloth is the output of a Computer Vision algorithm which is the slowest part of the scheme, taking multiple time steps to send an updated signal. This way, a linear ``Backup'' Simulation-Oriented Model (B-SOM) can simulate the cloth with the obtained control signal on each time step and update the initial conditions even if there is no new real feedback on that step. Once the feedback signal is processed, the state of this B-SOM is updated accordingly.

Both linear models (COM and B-SOM) follow the equations presented in Section~\ref{sec:proposed_solution}, but they can have different sizes. The nonlinear model simulating the real cloth is the one presented in \cite{franco_validation}, where a dynamic validation between their model and real cloth is performed (errors lower than 6\%). Computationally speaking, this model employs finite elements to discretize the Lagrangian of the mechanical system (kinetic energy minus potential energy) but not the equations of motion. This way, Euler-Lagrange equations are obtained as a system of ordinary differential equations (ODE) instead of a partial differential equations (PDE), making integrations faster. Even then, computations are slower than required for a real-time application, and this nonlinear model cannot be used as COM, and is used only in simulation.

The nonlinear model can be discretized with a mesh of $10\times10$ nodes to achieve realistic behavior \cite{franco_validation}. For the linear models, it was found that a mesh of $4\times4$ was enough for the considered trajectory tracking problem, as shown in Section~\ref{sec:results}. The proposed solution to be able to use different mesh sizes together is to make the smaller ones be sub-meshes of the larger ones. This can be done in square meshes when the side sizes $n$ follow $(n_L-1)=p(n_S-1)$ for some proportion $p\in\mathbb{Z}^+$. Knowing larger mesh sizes increase computational time, simulations were executed with sizes $n=10$ and 13 for the nonlinear model, and $n=4$ and 7 for the linear ones, testing cases where COM and B-SOM had different sizes or the same one. It can be checked how a mesh of $n=4$ can be extracted from one of $n=7$ and, in turn, this can be extracted from a larger mesh of $n=13$.

To evaluate the trajectory tracking performance, we define two Key Performance Indicators (KPIs). The first one is related to the tracking error, and can be obtained first computing the Root Mean Squared Error (RMSE) of the obtained output over time, which results in a 6-dimensional vector $\bm{e}$. Then this vector is split into the two corners, to compute the norm of each error and the final average, i.e.,
\begin{subequations}
\begin{align}
\bm{e} &= \text{RMSE}(\bm{y}-\bm{r}), \label{eq:RMSEe}  \\
\text{KPI}_1 &= \frac{1}{2} \left(\| \bm{e}^r \| + \| \bm{e}^l \| \right),  \label{eq:KPI1} \\
    \text{KPI}_2 & = \bar{\tau}. \label{eq:KPI2}
\end{align}
\label{eq:KPI}
\end{subequations}
\vspace{-6mm}

The second KPI is the average computational time per step $\tau$, without counting the time needed to simulate the nonlinear model. This is used to check if optimizations are completed within a time step.

\begin{figure}[H]
    \centering
    \includegraphics[width=0.98\linewidth]{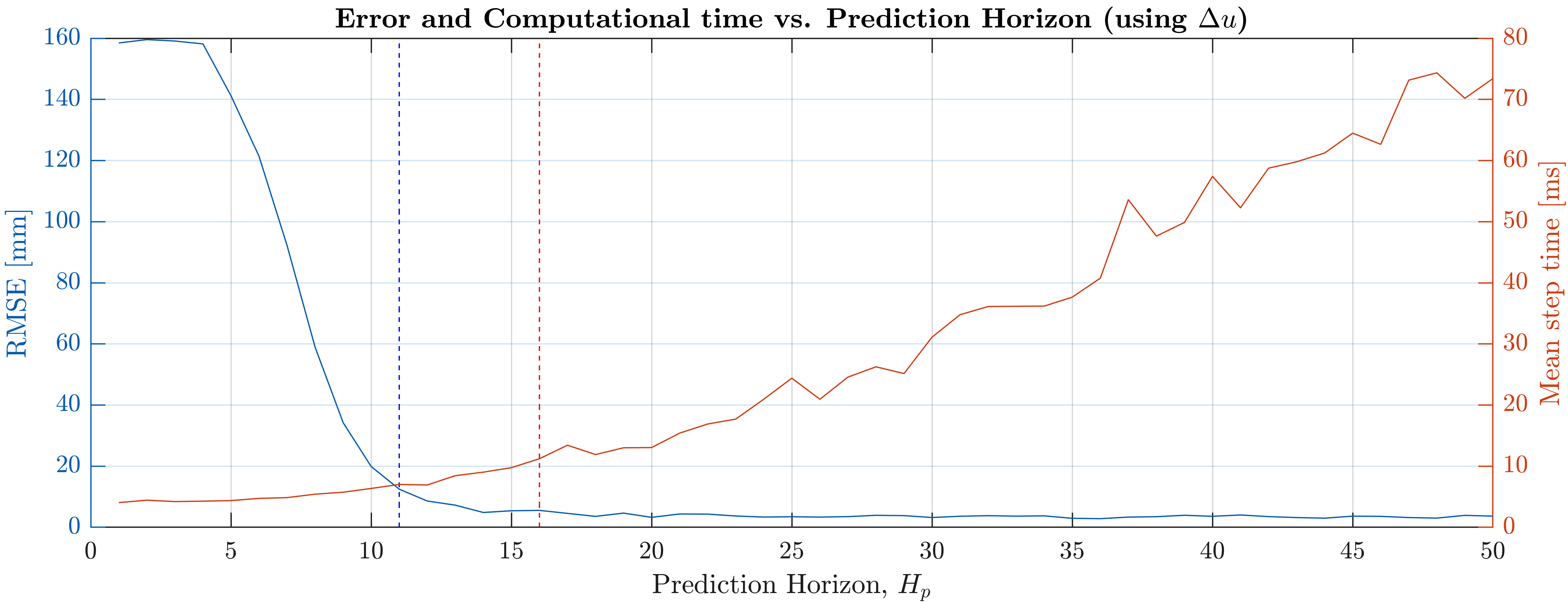}
    \caption{KPIs depending on $H_p$, for $T_s=10$ ms and $n=4$}
    \label{fig:hpanalysis_Du_020}
\end{figure}

These two KPIs were evaluated executing the same 3D trajectory with a range of different $H_p$ values to find a value that yielded low errors with times under $T_s$. The results for the case $T_s=0.01$ s and $n=4$ for both linear models are shown in Figure \ref{fig:hpanalysis_Du_020}. The vertical blue line represents the threshold where errors are 10\% larger than the minimum value found at $H_p=50$ steps. The red vertical line corresponds to the moment where times go over $T_s$. The window between these lines, $H_p\in[11,16]$ represents the range of possible values with optimal results. In the end, $H_p=15$ was chosen for experiments in these conditions.

It is worth noting $T_s=0.01$s is the most restrictive value among the tested ones. Larger time steps allow for larger prediction horizons. On the other hand, larger mesh sizes ($n=7$) are slower, meaning this analysis had to be done for each specific case, leading to similar results.

\subsection{Real Setup}
\label{sec:CS_real}
\begin{figure*}[t!]
    \centering
    \vspace{-1mm}
    \includegraphics[width=0.9\textwidth]{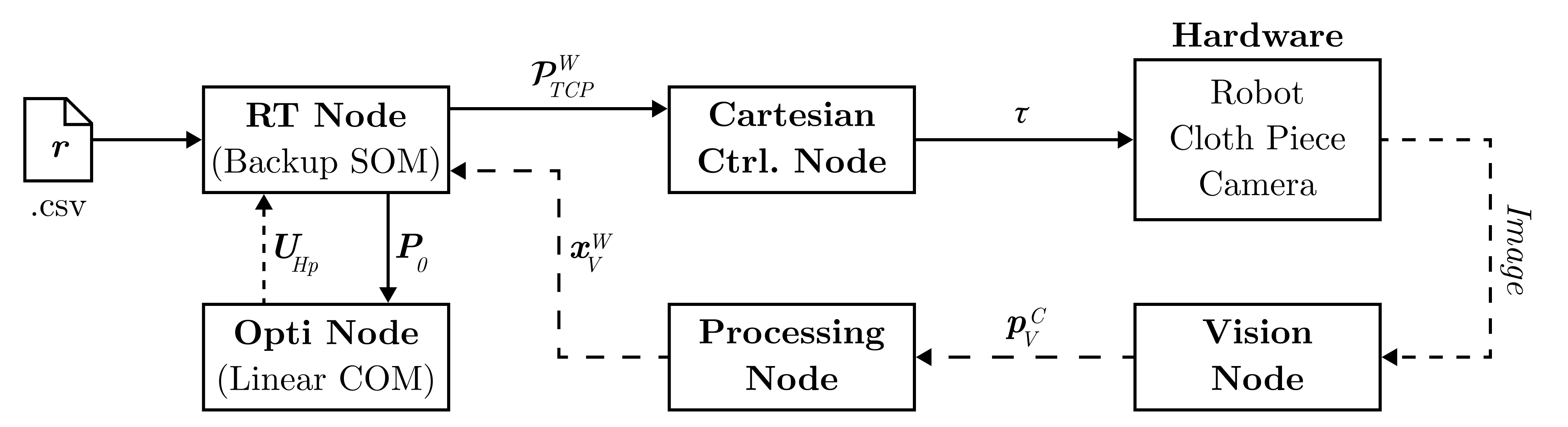}
    \caption{ROS Diagram of the implemented control scheme, with nodes and connections between them}
    \vspace{-3mm}
    \label{fig:ros_rt_scheme}
\end{figure*}

The final closed-loop control scheme was implemented using ROS Kinetic \cite{ros_kinetic}. Figure \ref{fig:ros_rt_scheme} shows the full diagram indicating the different nodes involved. The messages that are published at a fixed rate (one per time step) are marked with a solid line, and with dashed lines all which are not. The separation between dashes is proportional to how slow these messages are published, with the optimizer being the fastest among them (it can finish within one time step or take longer, depending on initial conditions). The Vision node used needs around $100$ms to process data and publish the mesh positions, thus its output and the output of the processing node are the slowest ones. The contents of each message are also specified, with $\bm{p}_V^C$ being the positions of all nodes expressed relative to the reference frame of the camera, $\bm{x}_V^W$ being the state vector obtained with Vision data in world base, and $\mathcal{P}_{TCP}^W$ is the pose of the TCP (position and quaternion) in global coordinates.

The RT Node processes the feedback data and updates the B-SOM state, the additional backup model needed to have updated data between real feedback samples, as discussed before. Every time step, the variable of initial parameters, $\bm{P}_0$, is updated and published to the Optimizer Node. The theoretical rate of the Opti Node is also set to $1/T_s$, but optimizations can take longer than that to complete. Once done, the full sequence of control inputs, $\bm{U}_{Hp}=\bm{u}^{s,*}$, is published and saved on the RT Node.

\begin{figure}[hb!]
    \centering
    \includegraphics[width=0.95\linewidth]{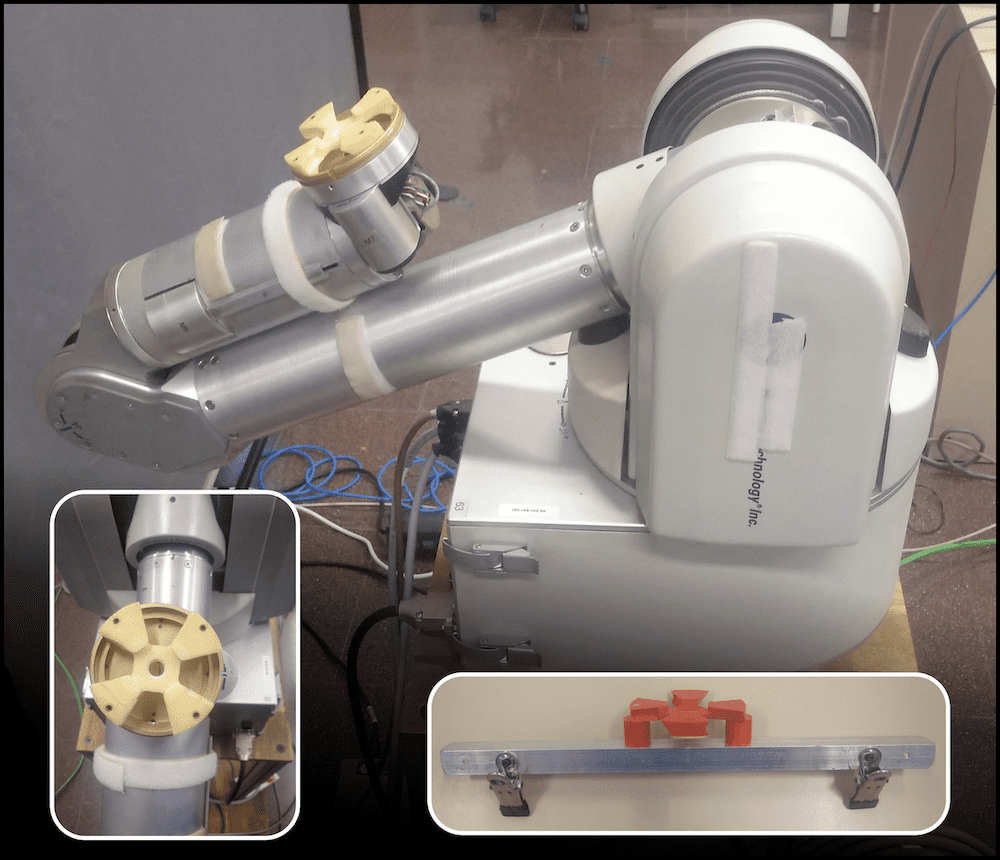}
    \caption{Picture of the WAM used in the real setup, with details of its end-effector and piece that connects to the cloth}
    \label{fig:realsetup_pieces}
\end{figure}

The Cartesian controller used to transform from a pose in Cartesian space to torques in joint space is the one developed in \cite{david_parent_cartesian_vic}, wrapped as a ROS node. The hardware includes a 7-DoF WAM robot from Barrett \cite{wam_7dof_dh}, used in all experiments with a custom gripper to hold the cloth by its upper corners (see Figure \ref{fig:realsetup_pieces}), and a Kinect camera to obtain RGB-D images as real feedback data.

These images were processed with the Vision Node \cite{miguel_arduengo_vision}, obtaining all node positions for a mesh of any given size. As mentioned, this process is the slowest step in the scheme, at rates around $0.1$s, creating the need of the B-SOM included in the RT Node.

The vector of positions needed some processing before closing the feedback loop. First, positions are expressed relative to the camera, so a change of base was performed. The camera reference frame can be obtained with available data, as we can obtain the pose of the end-effector from both references: using Forward Kinematics (FK) from the robot, and adding the known constant offset from the upper central point of the cloth that the camera is seeing. This can be written as
\begin{equation}
    T_C^W = T_E^W \cdot T_C^E = T_E^W \cdot \left( T_E^C \right)^{-1},
    \label{eq:4_tcw_tew_tec}
\end{equation}
where $W$ indicates the world/robot reference frame, $C$ the camera and $E$ the end-effector one. Transform $T_E^C$ is computed in a calibration process as an average inside a time window (set experimentally to 20 s), as the camera has a noisy output, with mesh positions varying slightly from frame to frame even with the cloth being completely still. After calibrating, $T_C^W$ is saved for the experiments.

Besides this change of reference, feedback data also had to be filtered. With the Vision Node being the slowest part of the system, waiting for the next sample to filter the current one would add a delay between four and ten steps, which was deemed too much and thus discarded. From the existing online filtering options using only current and past data points, the most commonly used ones are Moving Average (MA) filters \cite{moving_average_filters}, with several variants depending on the weights of the points considered. The disadvantage of using these filters is that their output corresponds to an average of the last several points, which in an actual movement means the filtered output lags behind the real position of the mesh, pulled back by the previous positions, adding a form of delay.

The considered variants were Simple (SMA), Weighted (WMA) and Exponential (EMA). The same set of experiments was executed for all three variants and different weight values, as well as using no filter. The best results (KPI$_1<6$~cm, KPI$_2<T_s$) were obtained using EMA, i.e.,
\begin{equation}
    \bm{y}_f(k) = \alpha \bm{y}(k) + (1-\alpha) \bm{y}_f(k-1),
\label{eq:4_filtereq}
\end{equation}
with $\alpha=0.66$, also obtained empirically. Therefore, this was the selected filter for all experiments.

This filtered feedback data is sent back to the MPC node. To avoid spurious and unreliable data affecting the system, a final filtering step discards data points with delays or distances to the simulated B-SOM over certain thresholds. In the real implementation, the B-SOM satisfies another important function when real data is received. The incoming sample can be received on the RT node with a significant delay with respect to when it was captured ($t_c$). Setting this data directly as the current initial state can lead to unsuccessful tracking, hence the need to update it from
$t_c$ to current time. This is done with another instance of the B-SOM simulating the required steps with a history of previously applied control signals until the current step, yielding $\bm{x}_V$. Finally, the B-SOM state vector $\bm{x}_{SOM}$ is updated with the new sample with
\begin{equation}
    \bm{x}_{SOM} \gets W_V \bm{x}_{V} + (1-W_V) \bm{x}_{SOM},
    \label{eq:4_xsom_update}
\end{equation}
where $W_V$ is the Vision weight, analogous to an observer gain, added to reduce the effect of the remaining noise to the MPC and obtain smoother evolutions.

As a summary of all the steps the captured Vision data must go through to close the loop, we show Algorithm \ref{alg:4_feedback}.

\begin{algorithm}[!h]
\caption{Closing the loop with Vision feedback data}
\label{alg:4_feedback}
\begin{algorithmic}[1]
\Require Camera publishing RGB-D images
\For {\textbf{each} image captured at time $t_c$}
\State Use Vision Node to get the mesh positions $\bm{p}(t_c)$
\State Apply EMA to obtain $\bm{p}_f(t_c)$
\State Order the node positions left to right, bottom to top
\State Apply a base change from camera to world reference
\State Add velocities to obtain full state vector
\State Publish mesh using acquisition time stamp, $\bm{x}_V(t_c)$
\State Receive mesh on MPC Node: callback function
\State Compute delay $\Delta t = t-t_c$
\If {$\Delta t > \Delta t_{max}$}
    \State Discard feedback data
    \State Exit callback function
\Else
    \State Update data using B-SOM $\ceil*{\Delta t / T_s}$ steps, \newline \mbox{\qquad\quad} get $\bm{x}_V(t)$
    \If {$\| \bm{x}_V(t)-\bm{x}_{SOM}(t) \| > \Delta d_{max}$}
        \State Discard feedback data
        \State Exit callback function
    \Else
        \State $\bm{x}_{SOM}(t) \gets W_V \bm{x}_{V} + (1-W_V) \bm{x}_{SOM}$
\EndIf
\EndIf
\State Exit callback function
\EndFor
\end{algorithmic}
\end{algorithm}

\section{RESULTS}
\label{sec:results}

All simulations were performed using CPU power only on an i7-8550U @1.80GHz with 8GB RAM using the optimization toolbox CasADi \cite{casadi} in MATLAB \cite{matlab}. All real experiments were executed with the described setup, using ROS Kinetic (written in C++), the same CasADi toolbox and the Eigen library \cite{eigen_lib}. All executions were done with a square piece of cloth of $30\times30$ cm.

\subsection{Model Validation}
The linear model presented in \eqref{eq:state_space_model} has seven parameters to tune in order for the linear model to follow the same behavior as a real cloth: spring stiffness $k_x,k_y,k_z$, damping $b_x,b_y,b_z$ and the $\Delta l_{0z}$ length previously introduced. It was found that these parameters depend on mesh size and sampling time $T_s$. Therefore, a combination of parameters had to be found for each tested situation.

First, we gathered data of real cloth evolutions (open-loop executions, without MPC) following a set of trajectories, to ensure movement in all directions of space. After being filtered and regularized in time, these evolutions were used to find the parameters of the linear model via Reinforcement Learning (RL). Concretely, the Relative Entropy Policy Search (REPS) algorithm was used \cite{reps_paper}, penalizing the squared errors in node positions for the entire mesh, but with higher weights for the lower corner nodes, as they will be the output to be tracked.

\begin{figure}[H]
    \centering
    \includegraphics[width=\linewidth]{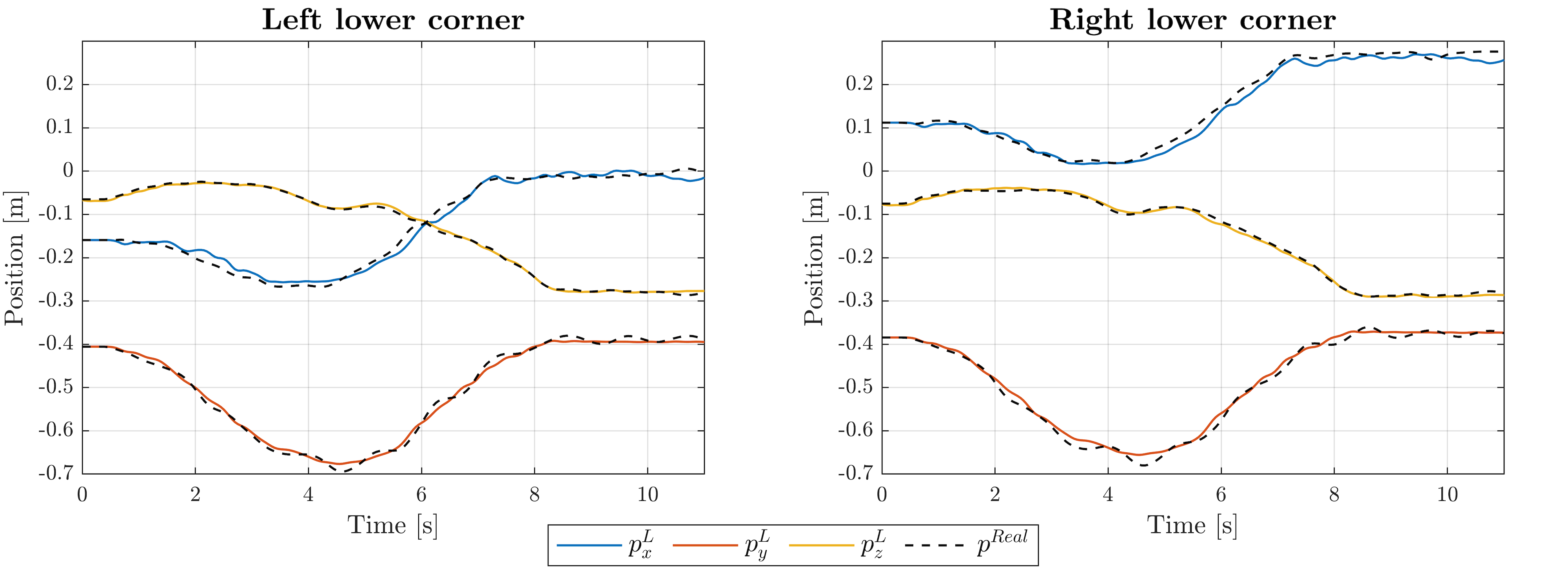}
    \caption{Lower corner evolutions of the learnt linear model ($n=4,\ T_s=15$~ms) and the data gathered from the real cloth (dashed line)}
    \label{fig:learn_model_evo}
\end{figure}

Learning experiments were conducted to learn the model parameters for square meshes of sizes $n=4$ and $7$, and for time steps of $T_s=10,\ 15,\ 20$ and $25$ ms, obtaining results for all eight combinations. Figure \ref{fig:learn_model_evo} shows the evolutions of the lower corner positions for the learnt linear model ($n=4,\ T_s=15$~ms) and the real cloth (dashed black line). We can see how both evolutions have the same behavior, with a final RMSE of 1.5 cm.

Even though errors increase with $n=7$ and larger time steps due to small oscillations that try to capture the nonlinear behavior of the real cloth, the errors are always within the same order of magnitude, and the lower corner evolutions of the linear model have the same behavior as the real ones. Therefore, the model in \eqref{eq:state_space_model} is validated with a real cloth, with parameter combinations for all studied cases.

\subsection{Controller Structure and Tuning}
Given the amount of parameters and weights present in the predictive controller, REPS was also used in closed-loop simulations to find its optimal structure and tuning. All experiments were made to learn the weight values inside matrices $\bm{Q}$ and $\bm{R}$ depending on different conditions, with the following options:

\begin{itemize}
    \item Using the adaptive weight in \eqref{adaptive_Q} or not to obtain $\bm{Q}$.
    \item Minimizing the control signals $\bm{u}$ or the slew rates $\bm{\Delta u}$.
    \item Three different structures of matrices $\bm{Q}$ and $\bm{R}$: scalars times the identity ($q\bm{I},r\bm{I}$), with different weights per coordinate only for $\bm{Q}$ ($\bm{Q}_{xyz},r\bm{I}$), or with both having different weights per coordinate, but with the matrices being proportional to each other ($\bm{Q}_{xyz}=k\bm{R}_{xyz}$).
    \item A reward function only penalizing tracking RMSE or also adding a cost for computational times over the considered step time ($\mbox{TOV} = \max( \bar{\tau}/T_s-1,\ 0)$).
\end{itemize}

\begin{figure}[!h]
    \centering
    \includegraphics[width=\linewidth]{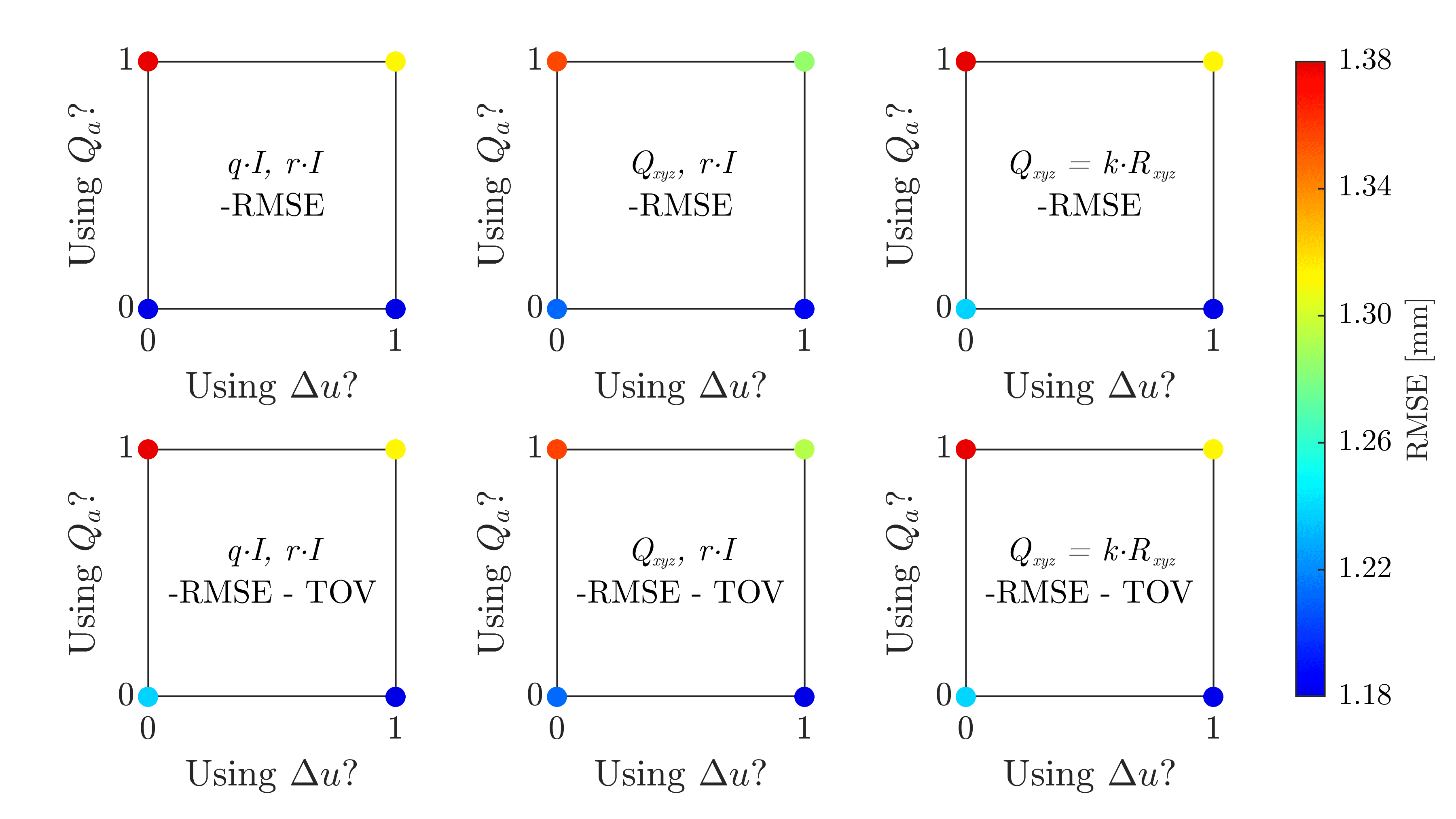}
    \caption{RMSE obtained with all the considered structure options}
    \label{fig:3_qaDu_RMSE}
\end{figure}

This results in a total of 24 learning experiments executed. We can organize their results in groups with the same weight structure and reward function, and make comparisons purely based on the use of $\bm{\Delta u}$ and $\bm{Q}_a$, as shown in Figure \nolinebreak \ref{fig:3_qaDu_RMSE}.

The worst results are always obtained using $\bm{Q}_a$ and $\bm{u}$ regardless of weight structure and reward function, and the opposite selection, with a constant $\bm{Q}$ and $\bm{\Delta u}$, yields the best outcome. This is the reason behind using $\bm{\Delta u}$ in \eqref{eq:optprob_onearm}.

Differences in computational time are minimal, with only a slight time increase when using $\bm{\Delta u}$. Knowing how time measurements are sensitive to memory and CPU usage, and obtaining roughly the same results with both reward functions, the second one, with TOV, was discarded for the following experiments.

The final tracking error is approximately the same regardless of the chosen weight structure (less than 0.5\% difference between best and worst), and the minimum errors are actually obtained with the first and simplest option. This means that the more complex structures, with weights depending on direction, do not adapt better to the trajectory leading to smaller errors. Experiments on two other trajectories confirmed this tendency, which, together with simplicity, pushed us to select the first structure, with just two weights, leaving only one degree of freedom to learn as the final controller tuning: the proportion between $\bm{Q}=q\bm{I}$ and $\bm{R}=r\bm{I}$.

Proceeding with more learning experiments, it was observed that the tracking errors decreased with low $\bm{R}/\bm{Q}$ ratios, until a certain threshold where $\bm{R}$ is too small and the system can unstabilize. This behavior was found to be consistent across all tested  trajectories and conditions, with the threshold only varying slightly between them.

This led us to set a unique tuning to guarantee the best performance in any given situation. The idea is getting as close as possible to the limit without crossing it for any case. A possible combination that works safely for all considered trajectories is $\bm{R}/\bm{Q}=0.2$ (e.g., $q=1,\ r=0.2$). It is worth noting that, even for the case where the threshold had the lowest ratio, using 0.2 increased the error less than 1 mm, which does not represent a great sacrifice in performance.

\subsection{Trajectory Tracking Results}


With the scheme successfully tested in simulation, real experiments were conducted with the conditions explained in Section~\ref{sec:CS_description}. To obtain the best trajectory tracking, an experimental analysis was conducted with two parts. The first one compared results with different linear model sizes in the same set of situations, while the second analyzed the effects of the remaining control parameters ($T_s,\ H_p,\ W_V$) on tracking performance.

\subsubsection{Linear model sizes}
\begin{figure}[ht!]
    \centering
    \includegraphics[width=\linewidth]{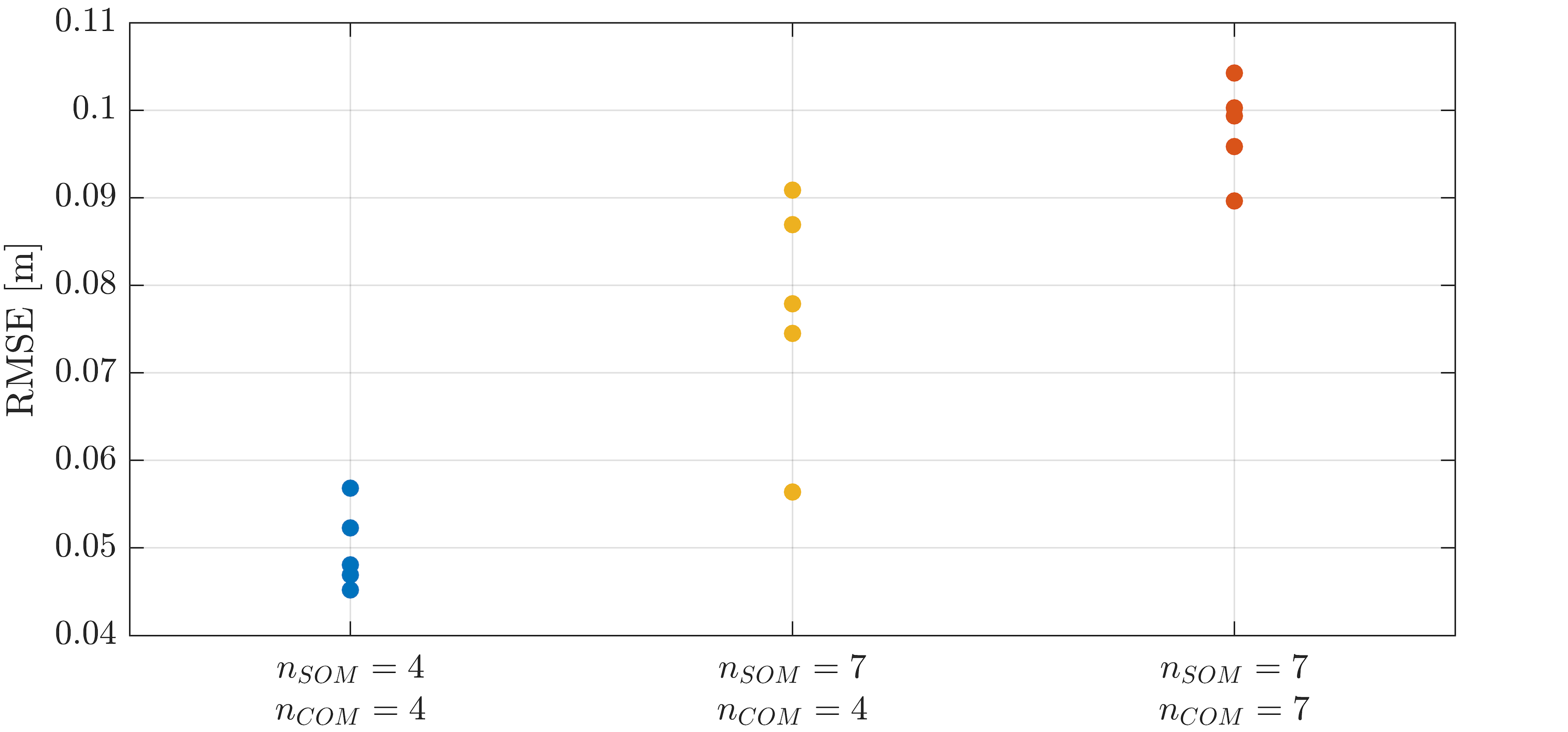}
    \caption{Obtained RMSE using different linear model sizes}
    \label{fig:5_3_mdlszs}
\end{figure}

Given that we have two available sizes and two linear models, and that the COM must be simpler or the same as the B-SOM, there are three possible combinations. Executing them in five different situations, we obtained the results shown in Figure~\ref{fig:5_3_mdlszs}. It is clear how larger mesh sizes yield worse results, meaning the increase in computational time is more important than the improvement in accuracy. Therefore, size $n=4$ was selected for both linear models in the following experiments.

\subsubsection{Control analysis}
\begin{figure}[ht!]
    \centering
    \includegraphics[width=\linewidth]{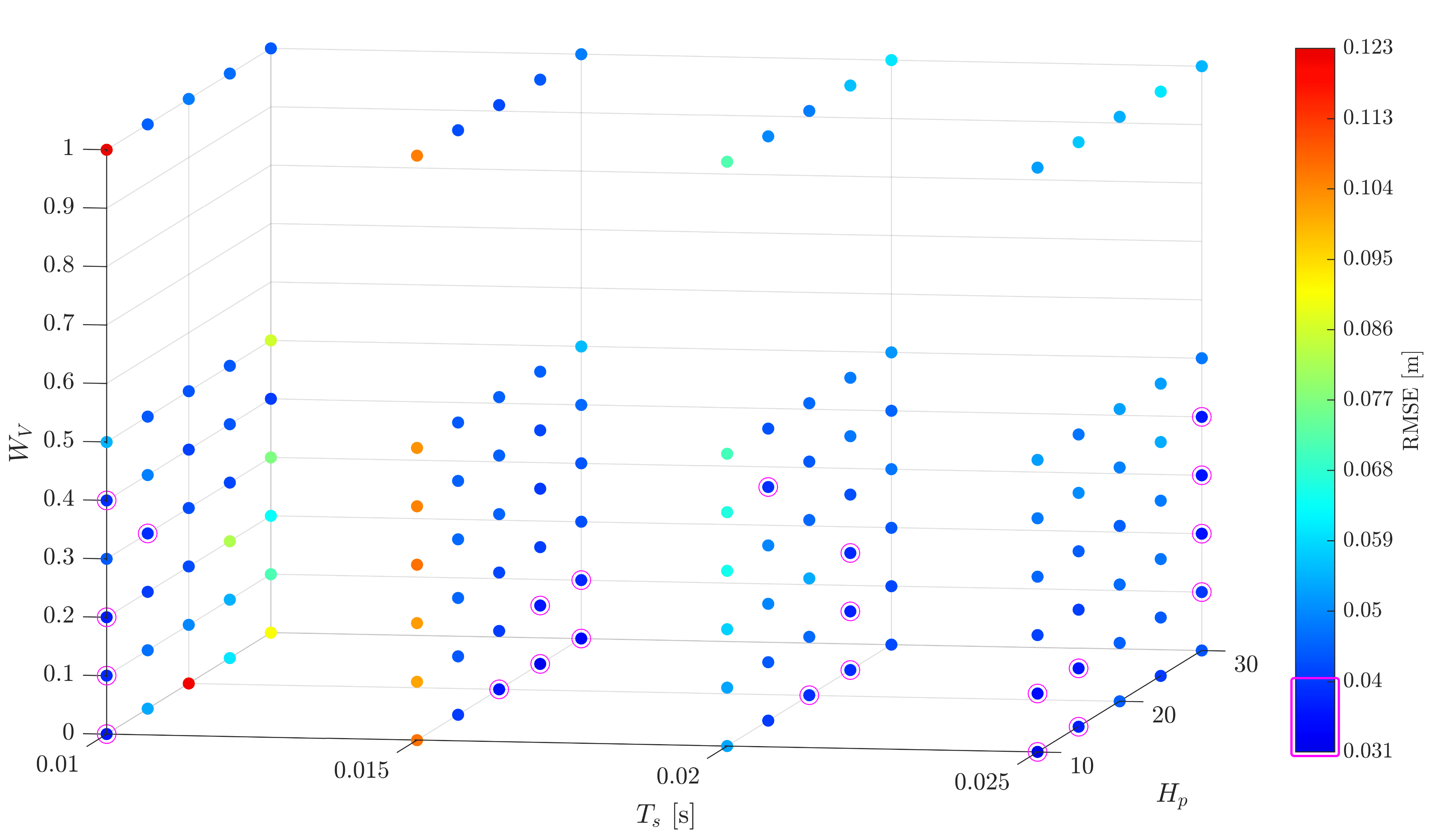}
    \caption{Results of all executed experiments depending on $T_s,\ H_p,\ W_V$}
    \label{fig:5_3_ctrl_analysis}
\end{figure}

The range of $T_s$ values analyzed comes from the obtained parameters through learning, being 10, 15, 20 and 25 ms. The prediction horizon $H_p$ can take any arbitrary ($\mathbb{Z}^+$) value, but thanks to the executed simulations, we have an indicative range with relatively low errors without increasing computational times over the limit. As the conditions in the real setup are different with regards to timing and feedback data, horizons were tested from 10 to 30 steps, in increases of 5 (10, 15, 20, 25 and 30). Finally, the weight $W_V$ was tested in increases of 10\% from 0 to 50\%. While higher values were also tested, the majority of combinations resulted in unsuccessful executions. This results in a total of 120 finished real-robot experiments under the same conditions except for these three parameters. After they were completed, however, some additional changes were made to test executions with higher $W_V$. They only worked after reducing the Cartesian controller gains and using a filter with $\alpha=0.5$, but all cases with $W_V=1$ (100\%) finished correctly. This makes a total of 140 experiments, with all their results shown in Figure~\ref{fig:5_3_ctrl_analysis}. Given the conditions for $W_V=1$ had to be different from the rest, and that the obtained results were consistently worse than their $W_V=0.5$ counterparts, no values of $W_V$ were tested in between to keep them separated and focus on the 120 executed with the same conditions. It is worth noting that the worst result within these 120 experiments had an RMSE of 12.1 cm, thus adding the additional experiments does not change the color scale significantly. We have also marked all results within the best 10\% of errors, i.e., with an RMSE lower than 4 cm with a magenta circle.

With these results, we can clearly see how a low $T_s$ combined with a high $H_p$ does not produce correct tracking. During execution, these experiments produced several timeouts on the optimizer, trying to keep up with a very fast rate while predicting a lot of steps into the future. Other issues were detected during the execution of all experiments with $T_s=10$ ms and $W_V\geq0.3$, where feedback data was consistently being discarded either for it being too much into the past (could not be updated) or too distant to the simulated state of the B-SOM. This means that these experiments actually have an effect of the feedback data closer to $W_V=0$ than their actual value, as most of the time the real data was unusable. Furthermore, a smaller time step results in an increased difference between the rate at which the control signals are computed and sent to the robot and the rate at which the Vision node outputs feedback data, being around 10 times slower in these conditions. This makes computed errors less reliable, as there are fewer captured points to compare with the reference. With all this, even if we have some results with really good tracking errors using $T_s=10$ ms, it is clear that this sampling time is too fast for the majority of cases, and must be avoided to obtain the best possible tracking results.

For both $T_s=15$ ms and $T_s=20$ ms, a horizon of 10 steps is too short to track correctly, regardless of $W_V$. This effect disappears with $T_s=25$ ms, having optimal results with the shortest $H_p$ too, which means that it is a problem related to total prediction time and maximum allowed time for the optimizer.

We can also see a tendency of errors increasing with higher $W_V$, with some optimal results being obtained without considering the Vision feedback at all. This is a product of noise, present even after filtering. In the executed experiments, there were no strong rotations, sudden movements, offsets, nor other disturbances (e.g., wind or human actions), and the B-SOM always started in the exact same position as the real cloth, making its simulated evolution an accurate one without any added noise. Of course, $W_V=0$ means the control loop is not closed, and cannot be applied in a general scenario, where external forces or initial deviations can make the B-SOM state have an unreliable evolution, and the real feedback would have to correct its state. Unfortunately, with the current camera and Computer Vision algorithms, this comes at the cost of updating the initial state of the MPC with noisy data, which can increase optimization times. A general application of this scheme would need a camera with a faster refresh rate, more precision, not fixed in place to enable more movements and orientations without losing the cloth, and a fast and reliable algorithm to obtain an updated mesh.

All in all, even discarding cases with $T_s=10$ ms, $W_V=0$ and 1, and the discussed cases with $H_p=10$, there is no singular delimited region with optimal results, but a tendency towards longer prediction horizons as the time step increases and the MPC module has more time to compute. Even then, all the remaining cases, 65 different combinations, yielded errors lower than 5.5 cm, which is not far from the previously considered threshold, and an acceptable error considering the large range of options it includes, and the precision of the camera and Vision algorithms used, also in the order of centimeters.

\subsubsection{Final tracking performance}
\begin{figure}[!h]
    \centering
    \includegraphics[width=\linewidth]{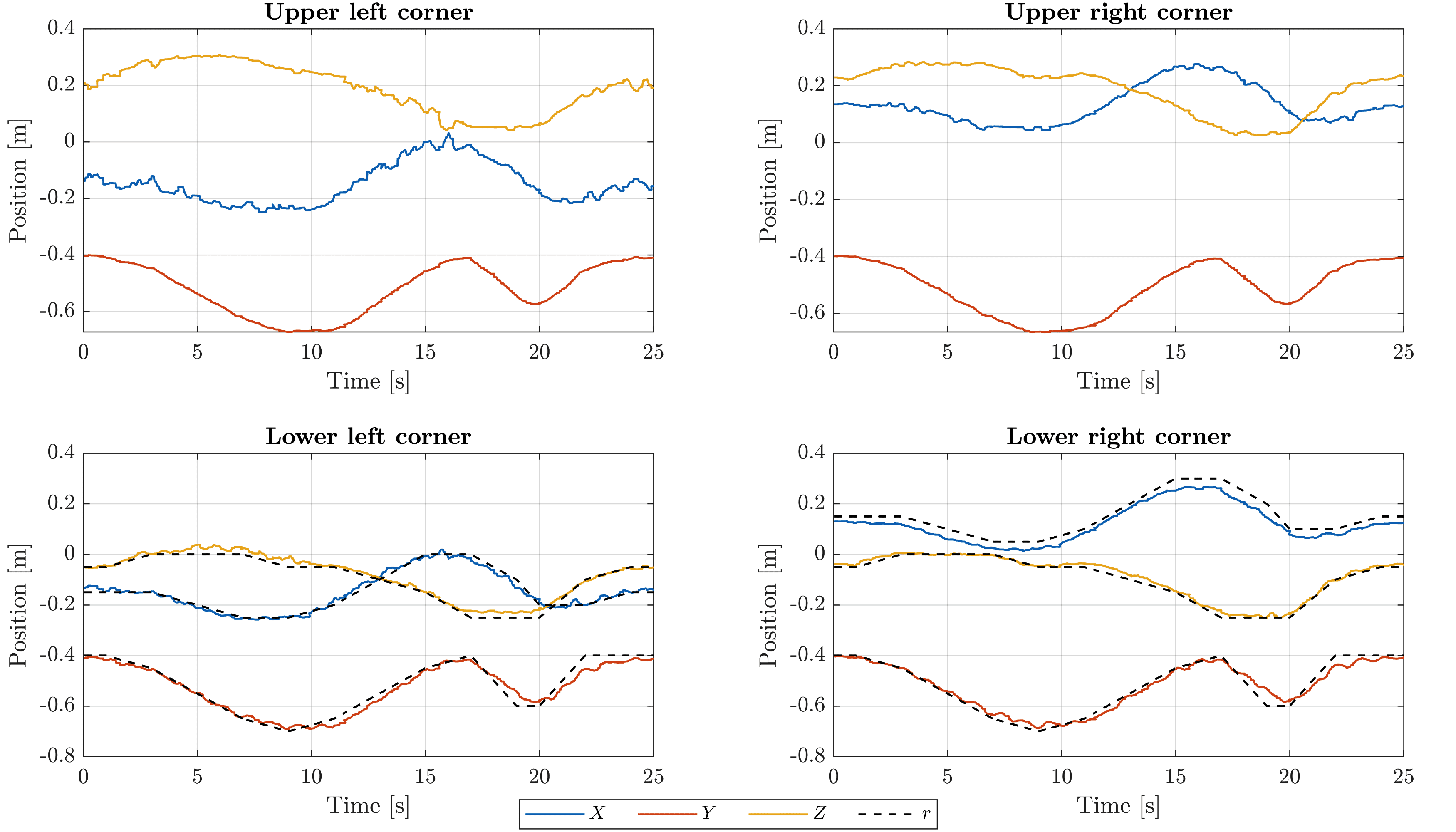}
    \caption{Cloth corner evolutions for $T_s=20$ ms, $H_p=25$, $W_V=0.2$}
    \label{fig:5_3_best_params}
\end{figure}

Choosing $T_s=20$ ms, $H_p=25$ and $W_V=0.2$ as an example of a combination that yielded optimal tracking, with an RMSE of only 3.8 cm, we can plot the evolutions of all corners, as shown in Figure~\ref{fig:5_3_best_params}. We can see how some noise persists, but the reference is tracked accurately.

\subsection{Experiments with Disturbances}
Two different cases were studied: blocking the camera and applying forces to the robot arm during execution.

\subsubsection{Blocking the camera}
\begin{figure}[ht!]
    \centering
    \includegraphics[width=\linewidth]{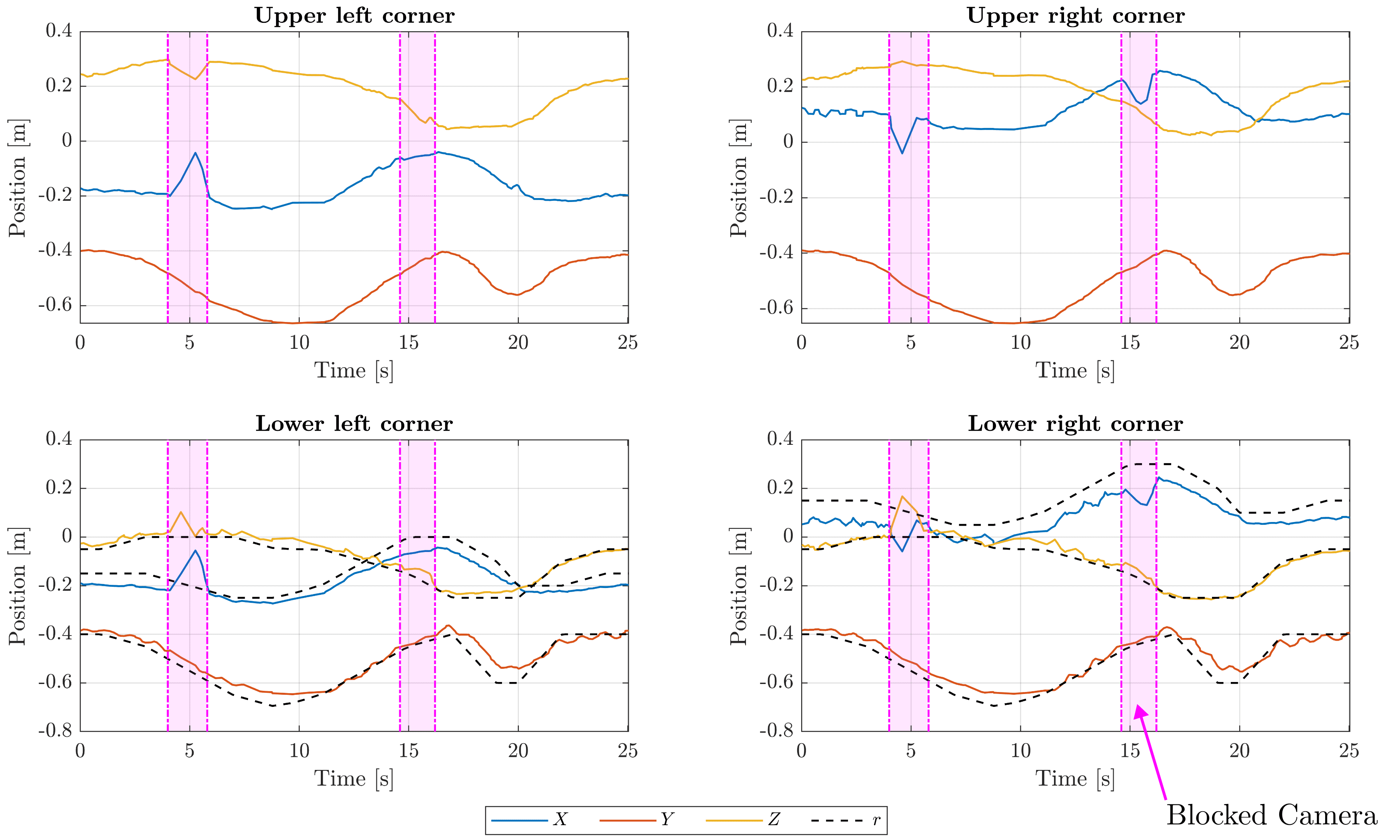}
    \vspace{3mm}
    \includegraphics[width=0.85\linewidth]{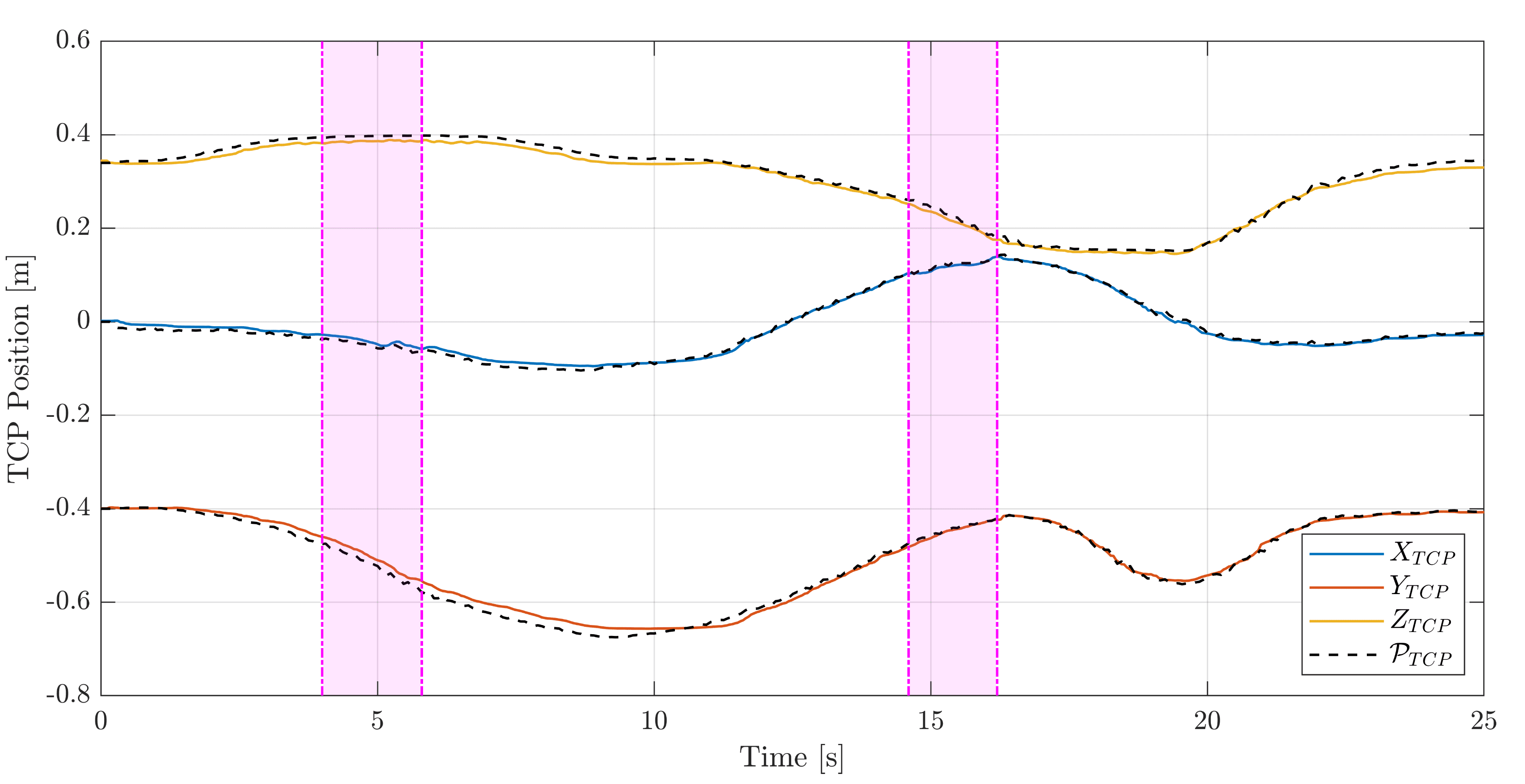}
    \caption{Cloth corners and TCP evolutions with camera blocking}
    \vspace{-3mm}
    \label{fig:5_4_3_dist1}
\end{figure}

During execution, a human walked between the camera and the cloth, covering its view for about two seconds on two occasions. The obtained results are shown in Figure~\ref{fig:5_4_3_dist1}, where shaded areas correspond to moments where the camera was blocked.

Sudden changes in captured mesh positions are caught before updating the state of the linear models and discarded, so even if we see them in the Vision data, they do not affect the controller. This is checked with the obtained TCP evolutions, smooth even during these moments. This can be achieved thanks to the B-SOM inside the controller, which simulates the real evolution of the cloth during the moments where the vision feedback cannot provide updated data.

\subsubsection{Human interaction}
The used Cartesian controller allows movements in the null space of the WAM without offering much resistance \cite{david_parent_cartesian_vic}. If a human moves the elbow joint, while theoretically it would not change the TCP pose, during a real experiment there are slight displacements and forces that are transmitted from the arm into the cloth, producing disturbances that are picked up by the camera.

\begin{figure}[ht!]
    \centering
    \includegraphics[width=\linewidth]{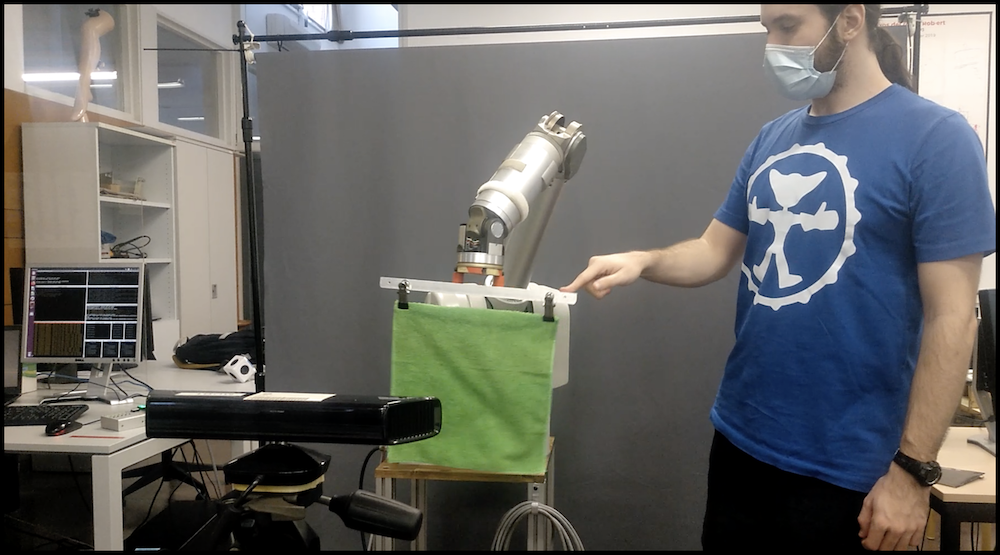}
    \caption{Disturbance created by a person poking the cloth}
    \label{fig:5_4_3_poking}
\end{figure}

In fact, the captured experiment started with a person walking in front of the camera as before, and after moving the elbow joint, the rigid piece connecting the upper corners of the cloth was pressed down on the right side, causing a slight rotation and a vibration when released. This situation can be seen in Figure~\ref{fig:5_4_3_poking}.

\begin{figure}[ht!]
    \centering
    \includegraphics[width=\linewidth]{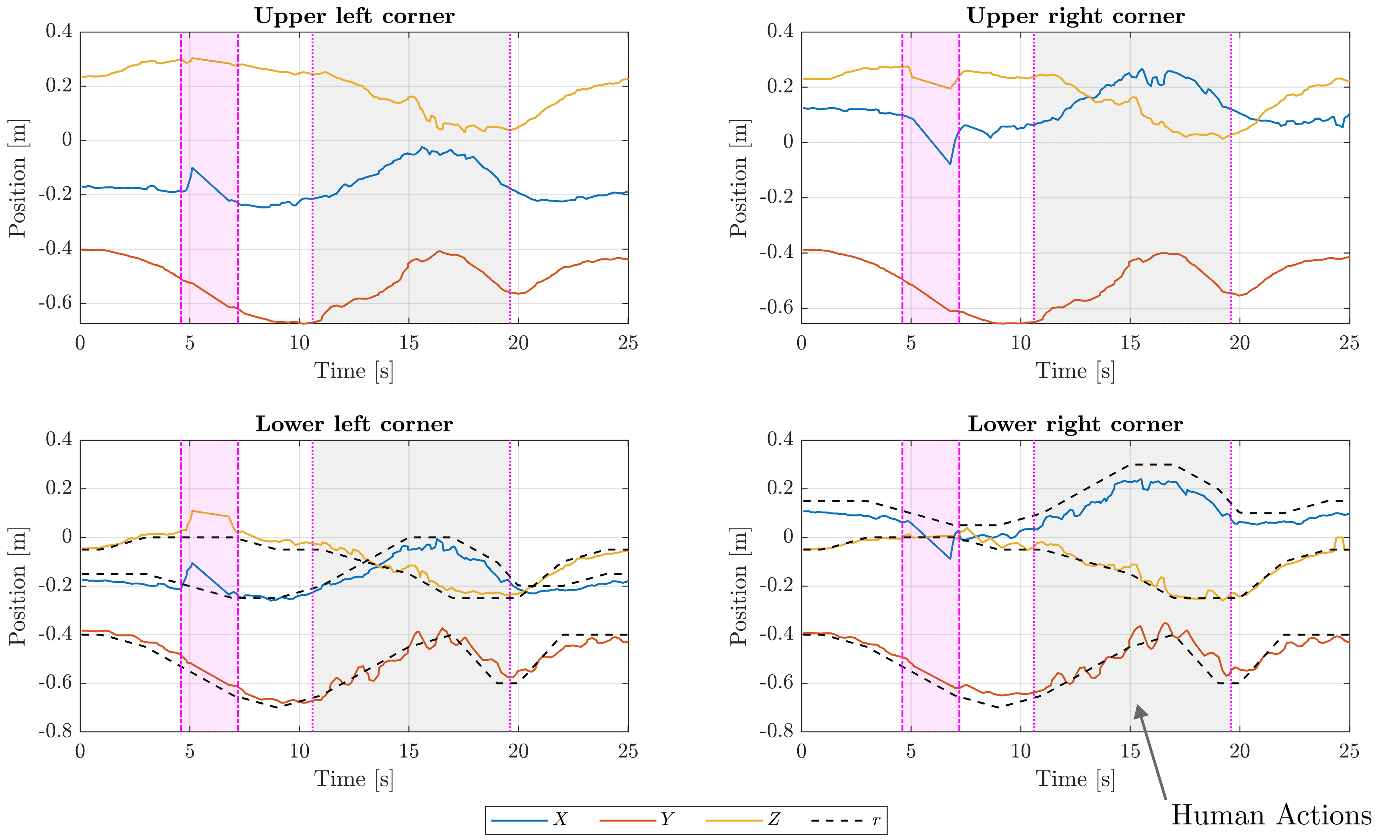}
    \vspace{3mm}
    \includegraphics[width=0.85\linewidth]{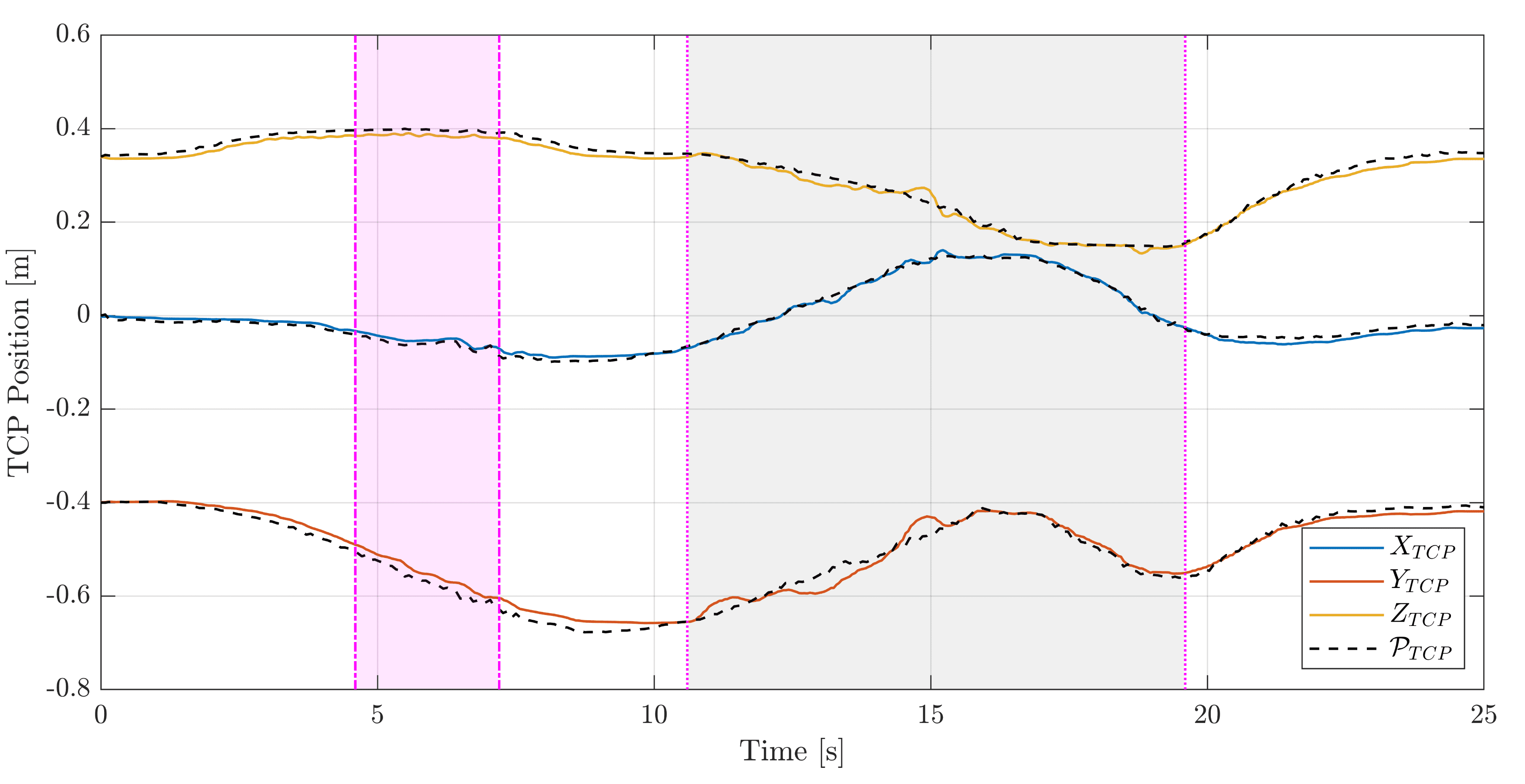}
    \caption{Cloth corners and TCP evolutions with human interaction}
    \vspace{-3mm}
    \label{fig:5_4_3_dist2}
\end{figure}

The results of this experiment are shown in Figure~\ref{fig:5_4_3_dist2}. The shaded region in magenta corresponds to the interval when the camera was blocked, as in the previous experiment, while the region shaded in gray corresponds to the time when a human agent was interacting with the robot arm.

Blocking the camera has the same effects as before. Moving the elbow of the arm (from around 11\,s to 15\,s) produces slight movements in the TCP, which are also visible in the upper corners, and are propagated to have a much greater effect on the lower corners due to the non-rigid nature of the cloth. When the rigid piece between the upper corners is pressed and released around the 15\,s mark, we can see how it creates a sudden disturbance, very clear in the right upper corner, which adds to the oscillations already present in the lower corners. Even in these situations, however, we can see how the reference trajectory is tracked correctly.

With these results, we can say that the developed implementation has proven to work in demanding conditions, and even under the effects of an external agent creating disturbances while keeping a good trajectory tracking performance. These last two experiments yielded RMSEs of 5.5 and 5.6~cm, respectively, once the incorrect data obtained when the camera was partially obstructed (magenta regions) had been removed.

\subsection{Error Analysis}
A final set of experiments was conducted with a completely new and different trajectory, which was also printed in real scale to record videos showing the obtained tracking behavior. One of these videos can be found in the supplementary material. These experiments were conducted to test the control scheme in situations outside the ones used for testing and obtaining the optimal parameters, and to do an analysis of all the possible error sources, quantifying each one.

From the components of the used control scheme, shown in Figure \ref{fig:ros_rt_scheme}, the ones that can introduce significant errors are i) the MPC, ii) the Cartesian Controller, iii) the physical robot, and iv) the Vision feedback, both from the actual camera limitations and the processing algorithm.

\begin{figure*}[ht!]
    \centering
    \vspace{-1mm}
    \includegraphics[width=0.96\textwidth]{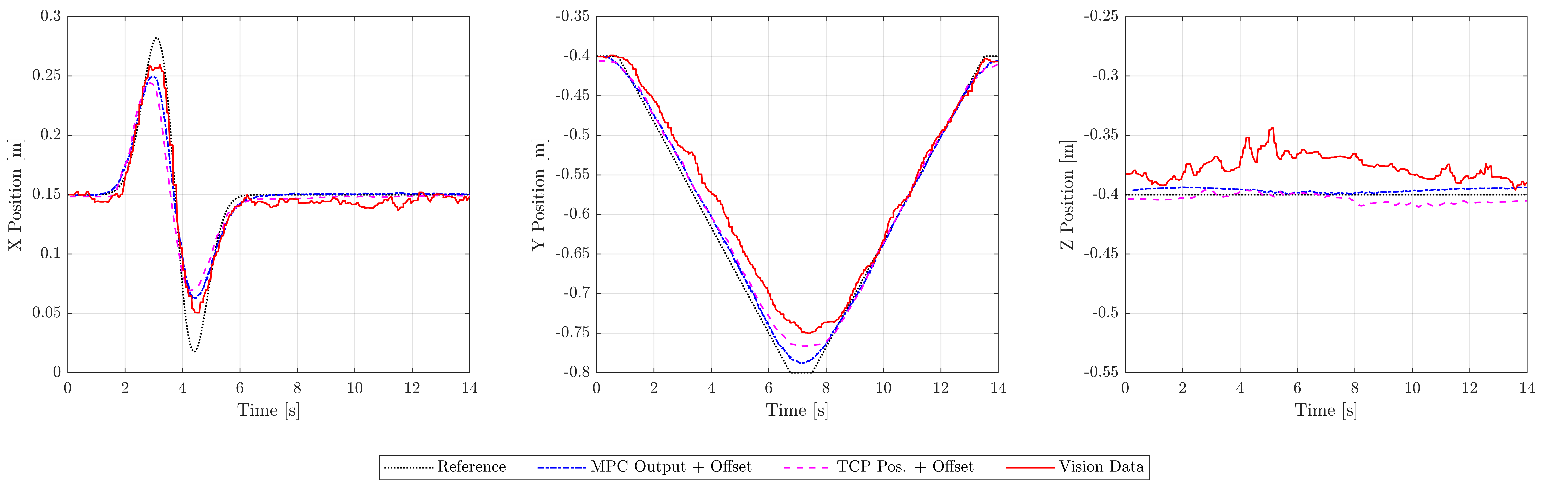}
    \caption{Obtained trajectories of all relevant signals in the control scheme, showing the different error sources}
    \vspace{-3mm}
    \label{fig:new_exps_err}
\end{figure*}

To evaluate the errors introduced by each of these components, we can compare the following data: i) the input reference, which is a sequence of lower corner positions, ii) the MPC output, a TCP pose from which we can obtain the corresponding desired TCP positions, iii) the actual position of the TCP at all times obtained via IK, and finally, iv) real Vision data, processed to obtain positions of the entire mesh every instant, and the lower corners in particular. To compare TCP (ii \& iii) and lower corner positions (i \& iv), we can add a fixed offset of the initial distance between them. This is an approximation, as the cloth is flexible and their relative position can change, but these differences are reduced on slow trajectories, as the tested one.

The new trajectory consists in a sinusoidal movement in the X-axis while moving closer to the camera on a constant height, followed by a linear movement to return to the initial position. In Figure \ref{fig:new_exps_err}, we show a comparison of the four aforementioned signals for the lower right corner in one execution.

The MPC output (plus a fixed offset) follows the reference trajectory closely in situations of constant velocity, but it tends to smooth out changes in direction. Part of this behavior can be explained due to its predictive nature, seeing future points after the direction change, and also given the flexible nature of the linear cloth model (COM), which will continue going in the previous direction due to inertia for a short time after the change. Even without taking these factors into account, the worst error is of 4.8~cm, and happens in one of the sharp changes in X. The RMSE between these two trajectories is of only 1.7~cm.

Additionally, we can compare the errors due to orientation, given that the reference trajectory has the TCP pointing straight down. Of course, we cannot say a change in rotation produces an error of the angle times the entire length of the cloth, as it is a non-rigid object, but we have a rigid link between the TCP and the cloth of 9~cm that can produce some position error due to changes in orientation. In this case, the MPC output follows the reference almost perfectly, with maximum errors of 0.1~mm in position due to orientation errors.

Next, we can obtain the error between the desired TCP pose and the real one, also saved throughout the experiment. In Figure \ref{fig:new_exps_err}, we can see how these two trajectories (blue and magenta) are close, with some discrepancies in the Z-axis and especially on the change of direction in the Y-axis. The maximum position error between the two is of 2.6~cm, with an RMSE of 1.7~cm. In the plots, we can see how in the most critical points (turns), the two mentioned errors are additive, meaning the actual TCP position is even further to the input reference. With regards to changes in position due to orientation errors, here they are more notable, with a maximum of 1~cm and an RMSE of 5~mm.

Finally, the last signal corresponds to the processed data from the Vision algorithm. While we could expect this signal to deviate from the TCP evolutions due to the flexible nature of the cloth piece, ideally being between the last signals and the input reference, here we are also adding the errors produced by the camera itself, with its maximum resolution and frame rate, and by the subsequent algorithm to detect and extract the positions of all the mesh nodes, and as we can see in Figure \ref{fig:new_exps_err} (red line), as in all previous experiments, the obtained data is very noisy. Besides that, it is clear how it is also unreliable in the Z-axis, as the TCP does not change in this direction, the cloth is always extended vertically, but we get an evolution with constant changes. This difference being the most severe in this axis can be due to the table placed directly under the cloth with the reference trajectory, to serve as a visual indicator for the videos, which might affect the detection process. In the X and Y axes, we can see how the reference is tracked correctly, with the largest deviations being in the direction changes. While in X the trajectory seems to be between the TCP and reference, in Y we always have it further from the reference, adding onto the previous errors.

\begin{figure}[ht!]
    \centering
    \includegraphics[width=\linewidth]{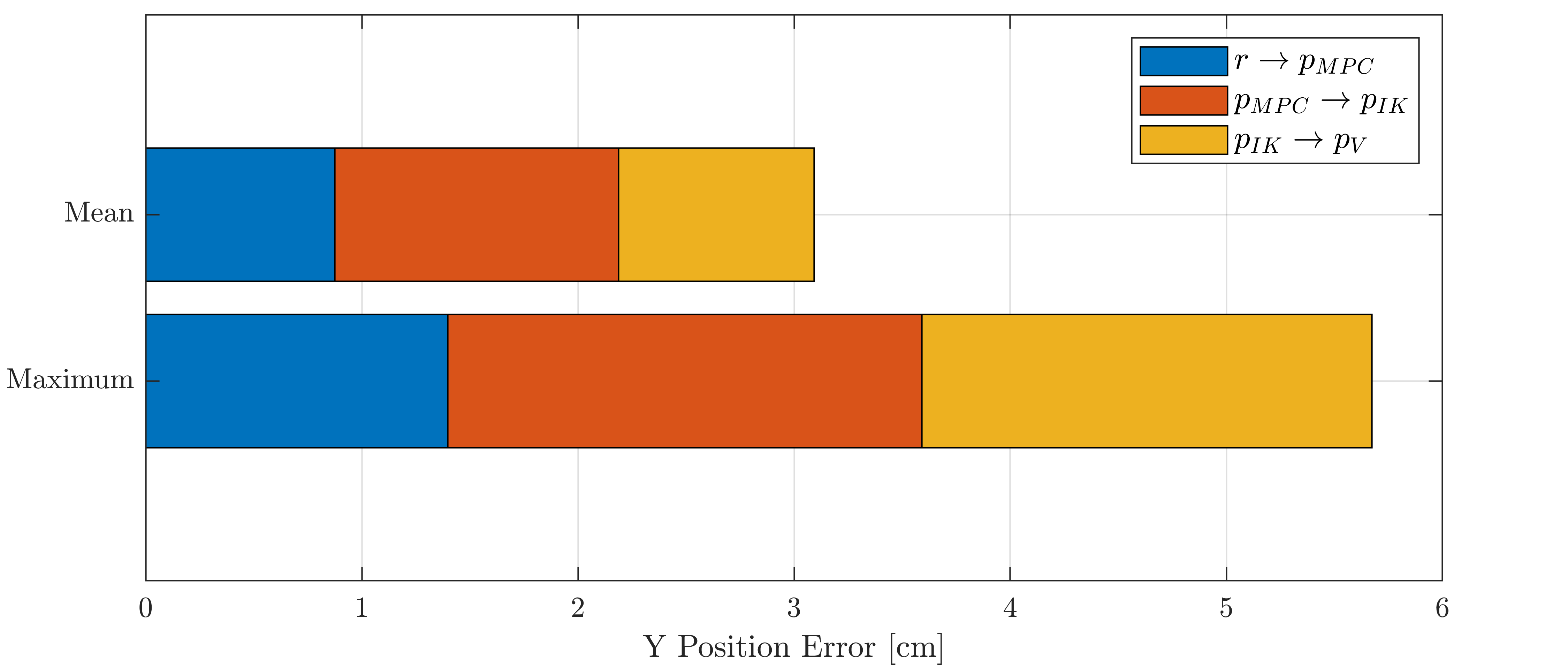}
    \caption{Mean and maximum errors in the Y-axis of the right lower corner, and their distribution in different sources}
    \label{fig:new_exps_ydist}
\end{figure}

The final RMSE of the experiment (from reference to Vision data) is of 4.1~cm, but the error is not constant or distributed evenly, as it reaches maximums of 4.4, 5.7 and 5.8~cm in X, Y and Z, respectively. While these errors can seem large at first, we must point out that they are the result of adding several sources together, and we do not have the exact position of the lower corners without camera noise and error, making difficult the isolation of the error due to the predictive controller. For example, for the worst case in the Y-axis, the difference between reference and MPC output is 1.4~cm, as shown in Figure \ref{fig:new_exps_ydist} (in blue), while the remaining 4.3~cm come from other sources (Cartesian controller and physical actuators in red, and not predicted cloth behaviors and Vision error together in yellow). Additionally, this difference is a conservative measure, as we are adding a constant offset to the TCP position and not accounting for the inertia and non-rigid behavior of the cloth, which, in theory, would help make the real evolution be closer to the input reference. We can see how this is not the case here, but as another example, in the X-axis, the captured evolution is closer to the reference than the MPC output. This is what we would expect in good tracking, as the MPC must consider these behaviors and correct them, and the TCP evolution must not be a perfect copy of the reference trajectory with a constant offset, as it would be with a rigid piece. In this case then, we can only remove the error between the desired and actual TCP pose (around 1~cm) and say that the remaining 3.4~cm of error are shared between MPC and Vision processing.

All in all, without separating error sources, we obtain values of KPI$_1$ similar to the ones obtained before, and this new analysis proves that the errors come from different parts of the control scheme and are added together, and how the Predictive Controller itself is not the main source of error.

\subsection{Comparison}

The obtained control strategy is suitable for dynamic cloth manipulation, but it should be compared with other existing techniques. Firstly, the simplest strategy one could think of is to use a classical controller such as a PID. However, it is not possible to use a PID because it cannot incorporate the dynamical model and the motion predictions.

Secondly, we can compare our proposal with existing MPC techniques applied to cloth manipulation. In \cite{SOTA_mpcforce}, the goal of the authors is to have position precision on the entire cloth. Hence, solving the optimization problem takes several minutes while we solve it in milliseconds, as we are only focused on the lower corners positions. Another existing technique for cloth manipulation is presented in \cite{berkeley_towel_25min}, where a PR2 robot folding towels is presented. However, the cloth manipulation is static or quasi-static, as in the majority of the existing literature. By contrast, our approach is to dynamically manipulate cloth by incorporating a dynamic cloth model in the control system. A different dynamic manipulation approach is presented in \cite{rishabh}, complementary to the one presented here. It tries to learn cloth movement without a previous model using RL. However, the learning requires thousands of simulated samples, and the policies generated for specific tasks would require further real-robot samples to transfer from simulation to reality.

\addtolength{\textheight}{-9.3cm}

\section{CONCLUSIONS}
\label{sec:conclusions}
In this paper, we have presented an MPC strategy to track the trajectories of the lower corners of a piece of cloth by only grasping the upper corners with one robot arm. To the best of the authors' knowledge, this is the first time that an MPC using a cloth model is used in real time for this kind of manipulation. Moreover, we have developed a linear cloth model that is shown to be an accurate approximation of real cloth and, therefore, is suitable to be used as a Control-Oriented Model in the MPC strategy.

Both the linear cloth model and the predictive controller have been improved via Reinforcement Learning, finding the parameters of the model in several scenarios to validate it against the behavior of a real cloth and enable its use inside the MPC, as well as learning the optimal structure and tuning of the controller.




Simulations showed accurate 3D trajectory tracking results, yielding mean errors in the order of millimeters for a cloth of 30$\times$30 cm, and solving the optimization problem in near real time.

After simulating, a full closed-loop control scheme was implemented in a real setup, including MPC, a Cartesian controller, and Vision feedback and processing. Working with a real environment raised additional issues, like noise and timing, which had to be addressed. The final implementation runs at a constant rate fixed by the chosen $T_s$, filters the output noise and has fail-saves against optimizations taking too long and vision data being too unreliable (with a B-SOM serving as a reference). Tracking errors increase with respect to the simulation values, to the order of centimeters, but they include new sources of error only present in the real setup, like sensor noise, actuator tolerances, and Cartesian controller inaccuracies. Even with disturbances, the errors (RMSE) are around the 5~cm mark, and an error analysis shows its principal source is not the Predictive Controller, even when considering a conservative approach.

The outcome of this paper can serve as a proof of concept of a new methodology for robotic cloth manipulation: using MPC with cloth dynamics prediction in real-time. A key aspect that allows for such real-time capability is the fact that, unlike other approaches \cite{SOTA_mpcforce}, we do not need to predict an accurate evolution for the entire cloth mesh, but rather few relevant points (e.g., lower corners plus the grasped points) can provide most of the information needed to achieve success in a task. There are, however, many aspects that can be further developed in the future:
 
\begin{itemize}
\setlength\itemsep{0em}
    \item While the cloth models proposed can be changed according to the user/task demands and situations, finding a general expression for the parameters of the linear model could improve the proposed method in this paper. We have seen how these parameters depend on both $T_s$ and mesh size $n$, and found the values for some combinations through learning, but an interesting step could be to try to find relations between these two conditions and the resulting parameters, and with enough time and data, even a model that can be used for any combination within a given range of values.
    \item Applying Online Learning. The found tuning and controller structure have proven to be the optimal ones for the tested cases, but for a more general application, a learning algorithm could use recent data during executions to adapt the parameters of the controller to their optimal values specifically for the task at hand.
    \item Change the Vision setup. This is probably the most important step to improve the setup, which includes using a faster and more precise camera, and a faster identification algorithm to obtain feedback data, but more importantly, changing from a situation where a single camera is fixed in place to another setup with either multiple cameras or mobile ones. This would allow for a much more varied range of trajectories and tasks that could be executed, and the control scheme could be tested to its maximum potential, without the Vision part being a bottleneck that limits manipulation speed.
\end{itemize}




\bibliographystyle{ieeetr}
\bibliography{root.bib}
\clearpage


\end{document}